\documentclass[12pt,smallextended]{svjour3}
\smartqed
\usepackage{lineno}
\modulolinenumbers[5]

\journalname{Journal of XXXX}
\usepackage{amssymb,amsmath,amsfonts}
\usepackage{epsfig}
\usepackage[bookmarksnumbered, colorlinks, urlcolor=blue, linkcolor=blue, citecolor=red, plainpages=false, pdfstartview=FitW]{hyperref}
\usepackage{mathptmx}
\usepackage{diagbox}
\usepackage{graphicx}
\usepackage{color}
\usepackage{caption}
\usepackage{wrapfig}
\usepackage{graphics}
\usepackage{enumerate}
\usepackage{amsxtra}
\usepackage{latexsym}
\usepackage{multirow}
\usepackage{slashbox}
\usepackage{subfigure}
\usepackage{epstopdf}
\usepackage{pdfpages}
\usepackage{algorithm}
\usepackage{algorithmic}

\usepackage{url}
\newcommand{\thmlist}{
\begin{list}{Step 1}
{\setlength{\leftmargin}{0.6 in}\setlength{\labelwidth} {0.5 in}}}

\newcommand{\alglist}{
\begin{list}{Step 1}
{\setlength{\leftmargin}{1.1 in} \setlength{\labelwidth}{1.0 in}}}

 \renewcommand{\proof} {\noindent {\bf Proof.} \quad}

\renewcommand{\subtitle}[1]{\color{blue}}

\begin{document}

\title{Continuation Newton methods with deflation techniques for global optimization problems}
\titlerunning{Continuation Newton methods for global optimization}
\author{Xin-long Luo\textsuperscript{$\ast$} \and Hang Xiao \and Sen Zhang}
\authorrunning{Luo \and Xiao \and Zhang}
\institute{
   Xin-long Luo, Corresponding author
   \at
   School of Artificial Intelligence,
   Beijing University of Posts and Telecommunications, P. O. Box 101,
   Xitucheng Road  No. 10, Haidian District, 100876, Beijing China,
   \email{luoxinlong@bupt.edu.cn}
   \and
   Hang Xiao
   \at
   School of Artificial Intelligence,
   Beijing University of Posts and Telecommunications, P. O. Box 101,
   Xitucheng Road  No. 10, Haidian District, 100876, Beijing China,
   \email{xiaohang0210@bupt.edu.cn}
   \and
   Sen Zhang
     \at
     School of Artificial Intelligence,
     Beijing University of Posts and Telecommunications, P. O. Box 101,
     Xitucheng Road  No. 10, Haidian District, 100876, Beijing China,
     \email{senzhang@bupt.edu.cn}
}

\date{Received: date / Accepted: date}
\maketitle

\begin{abstract}
The global minimum point of an optimization problem is of interest in
engineering fields and it is difficult to be found, especially for a nonconvex
large-scale optimization problem. In this article, we consider a new memetic
algorithm for this problem. That is to say, we use the continuation Newton
method with the deflation technique to find multiple stationary points of the
objective function and use those found stationary points as the initial seeds
of the evolutionary algorithm, other than the random initial seeds of the known
evolutionary algorithms. Meanwhile, in order to retain the usability of the
derivative-free method and the fast convergence of the gradient-based method,
we use the automatic differentiation technique to compute the gradient and
replace the Hessian matrix with its finite difference approximation. According
to our numerical experiments, this new algorithm works well for unconstrained
optimization problems and finds their global minima efficiently, in comparison
to the other representative global optimization methods such as the multi-start
methods (the built-in subroutine GlobalSearch.m of MATLAB R2021b, GLODS and
VRBBO), the branch-and-bound method (Couenne, a state-of-the-art open-source
solver for mixed integer nonlinear programming problems), and the
derivative-free algorithms (CMA-ES and MCS).
\end{abstract}


\vskip 2mm



\section{Introduction}

\vskip 2mm

In this article, we are mainly concerned with the global minimum
of the unconstrained optimization problem
\begin{align}
  \min_{x \in \Re^{n}} f(x),     \label{UNOPTF}
\end{align}
where $f: \; \Re^{n} \to \Re$ is a differentiable function except for a finite
number of points. When problem \eqref{UNOPTF} is nonconvex and large-scale, it
may have many local minimizers and it is a challenge to find its global minimum.
For problem \eqref{UNOPTF}, there are many popular global optimization methods,
such as the multi-start method
\cite{Boender1987,CM2015,DSDW2016,KN2022,RS2013,ULPG2007}, the branch-and-bound
method \cite{BLLMW2009,Couenne2020,Sahinidis2021,TS2005}, the genetic evolution
algorithm \cite{LW2001,MTK1999,Mitchell1996} and their memetic
algorithms \cite{Hart1994,MKCR2020,SGKZ2014,TRS2003}.

\vskip 2mm

For the multi-start method, we use the local  optimization method such as the
trust-region method \cite{CGT2000,DS1996,Yuan2015} or the line search method
\cite{NW1999,SY2006} to solve problem \eqref{UNOPTF} and find its local
minimum point $x^{\ast}$, which satisfies the following first-order
optimality condition
\begin{align}
   g(x^{\ast}) = \nabla f(x^{\ast}) = 0.   \label{NLEFX}
\end{align}
Since the implementation of the multi-start method is simple and successful for
many real-world problems such as the design of hybrid electric vehicles \cite{GM2007},
molecular conformal optimization problems \cite{AS1998,ACMX1992,LM2004} and
acoustic problems \cite{MHZHM2007}, many commercial global solvers are based on
the multi-start method such as GlobalSearch.m of MATLAB R2021b \cite{MATLAB}.
However, for the multi-start method, a random starting point could easily lie
in the basin of attraction of a previously found local minimum point \cite{RS2013}.
Hence, the multi-start method may find the same local minimum even if those
searches are initiated from points that are far apart. The other multi-start
method based on the deterministic procedure may have too many search branches
(the number of search regions is $2^{n\times l}$ at the $l$-th level iteration,
where $n$ is the dimension of the problem) \cite{CM2015,HN1999,Neumaier2000}.
Hence, the computational complexity is very high when the dimension of the
problem is large for the deterministic multi-start method.

\vskip 2mm

The evolutionary algorithm \cite{Moscato1989,SKM2018}
is another popular method due to its simplicity and attractive
nature-inspired interpretations, and is used by a broad community of engineers
and practitioners to solve the real-world problem such as finding the optimal location
for the active control actuator \cite{HSC2002}, the optimization of acoustic absorber
configuration \cite{XNF2004}, and the musical instrument \cite{Braden2005}. However,
for the evolutionary algorithm, since there is not a guaranteed convergence
and its computational time is very large for the large-scale problem, it
is difficult to find the global minimum accurately.

\vskip 2mm

In order to reduce the computational complexity of the multi-start method
and retain the advantage of the evolutionary algorithm escaping from the local
minima, we use the continuation Newton method \cite{AG2003,LXL2022,LX2021}
with the revised deflation technique \cite{BG1971} to find the multiple stationary
points of the objective function as the initial seeds of the quasi-genetic
algorithm, which are different from the random seeds of the known evolutionary
algorithms. The practical multi-start
method and the evolutionary algorithm are derivative-free methods. Namely,
the gradient of the objective function does not need to be provided by the user
and the derivative-free method \cite{CSV2009,RS2013B} is regarded as the black-box
method \cite{SK2017}.

\vskip 2mm

In order to retain the usability of the derivative-free method and the fast
convergence of the gradient-based method, we adopt the automatic differentiation
technique \cite{BPRS2017,GW2008,Neidinger2010,WV2013,WBB2014}
with the reverse mode to compute the gradient of the objective function and
replace the Hessian matrix with its finite difference approximation. In general,
the time that it takes to calculate the gradient of the objective function
by the automatic differentiation with the reverse mode is about a constant
multiple of the function cost. A rigorous analysis of all time costs associated
with the reverse method in general shows that the
cost is always less than four times the cost of the function evaluations
(pp. 83-86, \cite{GW2008}). Therefore, the computational complexity of the
gradient evaluated by the automatic differentiation with the reverse mode is far less than
that of the gradient estimated by the finite difference approximation or that of
the analytical gradient, which is about $n$ times the cost of the function
evaluations.

\vskip 2mm

Recently, Luo et al intensively investigate the continuation methods for
the eigenvalue problems \cite{LL2010,Luo2012}, linearly constrained optimization
problems \cite{LLS2022,LX2022}, the system of nonlinear equations
\cite{Luo2010,LXL2022,LX2021}, unconstrained optimization problems
\cite{Luo2005,LKLT2009,LXLZ2021}, the linear programming problem \cite{LY2022},
the linear complementarity problem \cite{LZX2022} and the nonlinear
equality-constrained optimization problems \cite{LXZ2023}. These continuation
methods are only applicable to find a solution of nonlinear equations. In this
paper, we need to confront multiple solutions of nonlinear equations. Namely, in
order to obtain the global minimum of the objective function, we need to find its
multiple stationary points. Therefore, we revise the deflation technique and
combine it with the continuation method to find multiple stationary
points of the objective function.

\vskip 2mm

The rest of this article is organized as follows. In the next section, we
convert the nonlinear system \eqref{NLEFX} into the continuation Newton
flow. Then, we use a continuation Newton method with the trust-region updating
strategy to follow the continuation Newton flow and find a stationary point of
the objective function. In Section 3, we revise the deflation technique to
construct a new nonlinear function $G_{k}(\cdot)$ such
that the zeros of $G_{k}(x)$ are the stationary points of the objective
function and exclude the found stationary points $x_{1}^{\ast}, \, \ldots, \,
x_{k}^{\ast}$ of the objective function. In Section 4, we give a quasi-genetic
algorithm to find a suboptimal point $x_{ag}^{\ast}$ by using the smallest
$L$ points $x_{1}^{\ast}, \, \ldots, \, x_{L}^{\ast}$ of all found stationary
points $x_{1}^{\ast}, \, \ldots, \, x_{\textbf{nsp}}^{\ast}$ as the evolutionary seeds.
After we obtain the suboptimal point $x_{ag}^{\ast}$ of the quasi-genetic
algorithm, we refine it by using it as the initial point of the continuation
Newton method to solve the system \eqref{NLEFX} of nonlinear equations and
obtain a better approximation of the global minimum point. In Section 5, we
give the descriptions of its implementation diagram and an example to
illustrate how to use this solver. In Section 6, some
promising numerical results of the proposed algorithm are reported, with
comparison to other representative global optimization methods such as the
multi-start method (the subroutine GlobalSearch.m of MATLAB R2021b, GLODS:
Global and Local Optimization using Direct Search \cite{CM2015}, and VRBBO:
Vienna Randomized Black Box Optimization \cite{KN2022}), the
branch-and-bound method (a state-of-the-art open-source
solver (Couenne) \cite{BLLMW2009,Couenne2020}), and the derivative-free
algorithm (the covariance matrix adaptation evolution strategy
(CMA-ES) \cite{Hansen2006,Hansen2010} and the multilevel coordinate search (MCS)
\cite{HN1999,Neumaier2000}). Finally, some discussions and conclusions are
also given in Section 7. $\|\cdot\|$ denotes the Euclidean vector norm or its
induced matrix norm throughout the paper.

\vskip 2mm

\section{Finding a stationary point with the continuation Newton method}
\label{SECCNM}

\vskip 2mm

For convenience, we restate the continuation Newton method with the trust-region
updating strategy \cite{LXL2022,LX2021},  which is used to find a stationary
point of $f(x)$, i.e. a root of nonlinear equations \eqref{NLEFX}.  We
consider the damped Newton method \cite{Kelley2003,Kelley2018,OR2000}
for finding a root of nonlinear equations \eqref{NLEFX} as follows:
\begin{align}
  x_{k+1} = x_{k} -  \alpha_{k} H(x_{k})^{-1} g(x_{k}), \label{NEWTON}
\end{align}
where $g(x)$ is the gradient of $f(x)$ and $H(x)$ is its Hessian matrix.

\vskip 2mm

If we regard $x_{k+1} = x(t_{k} + \alpha_{k})$, \; $x_{k} = x(t_{k})$ and let
$\alpha_{k} \to 0$, we obtain the following continuous Newton flow:
\begin{align}
  \frac{dx(t)}{dt} = - H(x)^{-1}g(x), \hskip 2mm  x(t_0) = x_0.
    \label{NEWTONFLOW}
\end{align}%
Actually, if we apply an iteration with the explicit Euler method
\cite{AP1998,SGT2003} for the continuous Newton flow \eqref{NEWTONFLOW},
we obtain the damped Newton method \eqref{NEWTON}. Since the Hessian
matrix $H(x)$ may be singular, we reformulate the continuous Newton flow
\eqref{NEWTONFLOW} as the following general formula:
\begin{align}
   H(x)\frac{dx(t)}{dt} = -g(x), \hskip 2mm  x(t_0) = x_0. \label{DAEFLOW}
\end{align}

\vskip 2mm

The continuous Newton flow \eqref{DAEFLOW} is an old method and can be backtracked
to Davidenko's work \cite{Davidenko1953} in 1953. After that, it was investigated
by Branin \cite{Branin1972}, Deuflhard et al \cite{DPR1975}, Tanabe \cite{Tanabe1979}
and Abbott \cite {Abbott1977} in 1970s, and applied to nonlinear boundary
problems by Axelsson and Sysala \cite{AS2015} recently. The continuous and even
growing interest in this method partially due to its global convergence as
stated by the following Proposition \ref{PRODAEFLOW}.

\vskip 2mm

\begin{proposition} (Branin \cite{Branin1972} and Tanabe \cite{Tanabe1979})
\label{PRODAEFLOW} Assume that $x(t)$ is a solution of the continuous Newton
flow \eqref{DAEFLOW}. Then, the energy function $E(x(t)) = \|g(x)\|^{2}$
converges to zero when $t \to \infty$. That is to say, for every limit point
$x^{\ast}$ of $x(t)$, it is also a solution of nonlinear equations \eqref{NLEFX}.
Furthermore, every element $g^{i} (x) \; (i = 1, \, 2, \ldots, \, n)$ of $g(x)$
has the same linear convergence rate and $x(t)$ can not converge to the
solution $x^{\ast}$ of nonlinear equations \eqref{NLEFX} on the finite interval
when the initial point $x_{0}$ is not a solution of nonlinear equations
\eqref{NLEFX}.
\end{proposition}
\proof For the proof of the first part i.e. the convergence $\lim_{t \to \infty}
g(x) = 0$, one can refer to references \cite{Branin1972,Tanabe1979}. The proof
of the second part can be found in references \cite{LXL2022,LX2021}. \qed

\vskip 2mm

Since the stiff concept of an ordinary differential equation (ODE) is a very
key role to understand the numerical method integrating an ODE trajectory, we
give its definition in Definition \ref{DEFSTIFF} for convenience.

\vskip 2mm

\begin{definition} (The definition of the stiff ODE \cite{HW1996,Lambert1973})
\label{DEFSTIFF}
An ordinary differential equation  $dx/dt = F(x)$ is called a stiff ODE, if
$\text{Re}(\lambda_{i}(x)) \le 0 \; (i = 1, \, 2,
\, \ldots, n)$ and the ratio $r(x) = \max_{i = 1, \, \ldots, n}
|\text{Re}(\lambda_{i}(x))|/\min_{i = 1, \, \ldots, n}
|\text{Re}(\lambda_{i}(x))|$ is very large, where $\text{Re}(\lambda(x))$
represents the real part of $\lambda(x)$ and $\lambda_{i}(x) \;
(i = 1, \, \ldots, n)$ are the eigenvalues of the Jacobian $F'(x)$.
\end{definition}

\vskip 2mm

\begin{remark}
The inverse $H(x)^{-1}$ of the Hessian matrix $H(x)$ can be regarded as the
pre-conditioner of $g(x)$ such that every solution element $x^{i}(t) \, (i = 1, \, 2,
\ldots, n)$ of the continuous Newton flow \eqref{NEWTONFLOW} has the roughly same
convergence rate, and it mitigates the stiffness of ODE \eqref{NEWTONFLOW}. This
property is very useful since it makes us adopt the explicit ODE method to follow
the trajectory of the Newton flow.
\end{remark}

\vskip 2mm

Actually, if we consider $g(x) = Ax$, where $A$ is an $n \times n$
nonsingular and symmetric matrix, from ODE \eqref{DAEFLOW},
we have
\begin{align}
  A \frac{dx(t)}{dt} = - A x, \; x(0) = x_{0}. \nonumber
\end{align}
Thus, we obtain
\begin{align}
  \frac{dx(t)}{dt} = - x, \; x(0) = x_{0}. \label{LINAPP}
\end{align}
From Definition \ref{DEFSTIFF}, we know that ODE \eqref{LINAPP} is a
non-stiff ODE. By integrating the linear ODE \eqref{LINAPP}, we obtain
\begin{align}
   x(t) = e^{-t} x_{0}.       \label{LINSOL}
\end{align}
From equation \eqref{LINSOL}, we know that every element
$x^{i}(t) \; (i = 1, \, 2, \, \ldots, n)$ of the solution $x(t)$
converges to zero linearly with the same rate when $t$ tends to infinity.

\vskip 2mm

When the Hessian matrix $H(x)$ is singular or nearly singular, ODE
\eqref{DAEFLOW} is the system of differential-algebraic equations (DAEs),
and its trajectory can not be efficiently followed by the general ODE methods
\cite{AP1998,HW1996,Jackiewicz2009} such as the backward differentiation
formulas (the built-in subroutine ode15s.m of MATLAB R2021b). Therefore, we
need to construct the special method to handle this problem.

\vskip 2mm

We expect that the new method has the global convergence as the homotopy
continuation method, and the fast convergence rate near the stationary point
$x^{\ast}$ as the merit-function-based method. In order to achieve the two
aims, we construct the special continuation Newton method with the new step
$\alpha_{k} = \Delta t_{k}/(1+\Delta t_{k})$
\cite{LXL2022,LY2022,LXLZ2021,LX2021,LX2022,LZX2022} and the time step
$\Delta t_{k}$ is adaptively adjusted by the trust-region updating strategy
\cite{Yuan1998,Yuan2015} to follow the trajectory of ODE \eqref{DAEFLOW}.

\vskip 2mm

Firstly, we apply the implicit Euler method \cite{SGT2003} to ODE
\eqref{DAEFLOW}. Then, we obtain
\begin{align}
    H(x_{k+1})\left(x_{k+1} - x_{k}\right) =  - {\Delta t_{k}} g(x_{k+1}).
  \label{IMEDAE}
\end{align}
The scheme \eqref{IMEDAE} is an implicit formula and it needs to solve a system
of nonlinear equations at every iteration. To avoid solving the system of
nonlinear equations, we replace the Hessian matrix $H(x_{k+1})$ with $H(x_{k})$,
and substitute $g(x_{k+1})$ with its linear approximation
$g(x_k) + H(x_k)(x_{k+1} - x_{k})$ in equation \eqref{IMEDAE}, respectively.
Thus, we obtain the following explicit continuation Newton method:
\begin{align}
    H_{k} s_{k}^{N} = - g(x_k),  \; s_{k} = \frac{\Delta t_k}{1+\Delta t_k}s_{k}^{N}, \;
    x_{k+1} = x_{k} + s_{k},  \label{CNM}
\end{align}
where $H_{k}$ equals $H(x_{k})$ or its approximation. For the symmetric indefinite
system \eqref{CNM}, it can be solved by the method of Aasen, which involves $n^{3}/3$
flops (see pp. 186-190 in \cite{GV2013} or reference \cite{BMR2022}).

\vskip 2mm

The continuation Newton method \eqref{CNM} is similar to the damped Newton
method \eqref{NEWTON} if we set $\alpha_{k} = \Delta t_k/(1+\Delta t_k)$ in
equation \eqref{CNM}. However, from the view of an ODE method, they are
different. The damped Newton method \eqref{NEWTON} is obtained by the explicit
Euler method applied to ODE \eqref{DAEFLOW}, and its time step $\alpha_k$ is
restricted by the numerical stability \cite{HW1996}. That is to say, for the
linear test function $g(x) = Ax$, from equation \eqref{LINAPP},
we know that its time step $\alpha_{k}$ is restricted by the stable region
$|1-\alpha_{k}| \le 1$. Therefore, the large time step $\alpha_{k}$ can not be
adopted in the steady-state phase.

\vskip 2mm

The continuation Newton method \eqref{CNM} is obtained by the implicit Euler
method and its linear approximation applied to ODE \eqref{DAEFLOW}, and its
time step $\Delta t_k$ is not restricted by the numerical stability for the
linear test equation \eqref{LINAPP}. Therefore, the large time step can be
adopted in the steady-state phase for the continuation Newton method
\eqref{CNM}, and it mimics the Newton method near the stationary point
$x^{\ast}$ such that it has the fast local convergence rate.

\vskip 2mm

The most of all, $\alpha_{k} = \Delta t_{k}/(\Delta t_{k} + 1)$
in equation \eqref{CNM} is favourable to adopt the trust-region updating strategy
for adaptively adjusting the time step $\Delta t_{k}$ such that the continuation
Newton method \eqref{CNM} accurately follows the trajectory of ODE the continuous Newton
flow in the transient-state phase and achieves the fast convergence rate near
the stationary point $x^{\ast}$.

\vskip 2mm

The main idea of the adaptive control step $\Delta t_k$ based on the trust-region
updating strategies \cite{Deuflhard2004,Higham1999,LL2010,LKLT2009,Luo2010,Luo2012}
can be described as follows. The time step $\Delta t_{k+1}$ will be enlarged
when the linear model $g(x_k) + H_{k}s_{k}$ approximates $g(x_{k}+s_{k})$
well, and $\Delta t_{k+1}$ will be reduced when $g(x_k) + H_{k}s_{k}$
approximates $g(x_{k}+s_{k})$ badly. Thus, by using the relation of equation \eqref{CNM},
we enlarge or reduce the time step $\Delta t_{k+1}$ at every iteration according
to the following ratio:
\begin{align}
  \rho_k = \frac{\|g(x_{k})\|-\|g(x_{k}+s_{k})\|}
  {\|g(x_{k})\| - \|g(x_{k})+H_{k}s_{k}\|}
  = \frac{1+\Delta t_k}{\Delta t_k} \,
  \frac{\|g(x_{k})\|-\|g(x_{k}+s_{k})\|}{\|g(x_k)\|}. \label{RHOK}
\end{align}

\vskip 2mm

When the gradient $g(x)$ is very nonlinear, a decrease in the gradient is rarely
found and $\Delta t_k$ becomes too small and even zero before a minimizer is
found. In such case, $\rho_k$ defined by equation \eqref{RHOK} goes to -$\infty$. To
overcome this problem, $\Delta t_k$ must be restricted from below by a tuning
small positive parameter $\overline{\Delta t} < 1$. Thus, by combining it with
equation \eqref{RHOK}, we give a following particular adjustment strategy:
\begin{align}
   \Delta t_{k+1} =
     \begin{cases}
          c_2 \Delta t_k, \; &\text{if} \; 0 \leq |1- \rho_k| \le \eta_1,
            \\
          c_1 \Delta t_k, \; &\text{if} \;  |1 - \rho_k| \ge \eta_2
          \; \text{and} \; \Delta t_{k} \ge \overline{\Delta t},
         \\
        \Delta t_k, \; &\text{others},
    \end{cases} \label{TSK1}
\end{align}
where the constants $c_1, \; c_2, \; \eta_1, \; \eta_2$ satisfy
$0 < c_1 < 1 \le c_2$ and $0 < \eta_1 < \eta_2 < 1$, and $0 < \overline{\Delta t} < 1$
is a small threshold value. When $\rho_{k} \ge \eta_{a}$, we accept the trial
step and set
\begin{align}
    x_{k+1} = x_{k} + s_{k},   \label{ACCPXK1}
\end{align}
where $s_{k}$ is solved by equation \eqref{CNM} and $\eta_{a}$ is
a small positive number. Otherwise, we discard the trial step and set
\begin{align}
     x_{k+1}  = x_{k}. \label{NOACXK1}
\end{align}

\vskip 2mm

\begin{remark}
This time-stepping selection based on the trust-region updating strategy
has some advantages compared to the traditional line search strategy \cite{Luo2005}.
If we use the line search strategy and the damped Newton method \eqref{NEWTON}
to track the trajectory $x(t)$ of ODE \eqref{DAEFLOW}, in order to achieve
the fast convergence rate in the steady-state phase, the time step $\alpha_{k}$
of the damped Newton method is tried from 1 and reduced by the half with many
times at every iteration. Since the linear model $g(x_{k}) + H_{k}s_{k}^{N}$ may
not approximate $g\left(x_{k}+s_{k}^{N}\right)$ well in the transient-state phase,
the time step $\alpha_{k}$ will be small. Consequently, the line search strategy
consumes the unnecessary trial steps in the transient-state phase. However, the
time-stepping selection method based on the trust-region updating strategy
\eqref{RHOK}-\eqref{TSK1} can overcome this shortcoming.
\end{remark}

\vskip 2mm

For a system of nonlinear equations, Hessian evaluations are too large if we
update the Hessian matrix $H(x_{k})$ at every iteration. In order to reduce
the computational time of Hessian evaluations, similarly to the adaptively
updating Jacobian technique \cite{LXLZ2021,LX2021}, we set $H_{k+1} = H_{k}$
when $g(x_{k}) + H_{k}s_{k}$ approximates $g(x_{k}+s_{k})$ well. Otherwise, we
update $H_{k+1} = H(x_{k+1})$. An effective updating strategy is give by
\begin{align}
     H_{k+1} = \begin{cases}
                 H_{k},  \; \text{if} \; |1- \rho_{k}| \le \eta_{1}, \\
                 H(x_{k+1}), \; \text{otherwise},
               \end{cases} \label{UPDJK1}
\end{align}
where $\rho_{k}$ is defined by equation \eqref{RHOK}. Thus, when $H_{k}$ performs
well, i.e. $|1-\rho_{k}| \le \eta_{1}$, according to the updating formula
\eqref{UPDJK1}, we set $H_{k+1} = H_{k}$ in equation \eqref{CNM}.

\vskip 2mm

According to the above discussions, we give the detailed descriptions of the
continuation Newton method and the trust-region updating strategy for finding a
stationary point of $f(x)$ (i.e. a root of nonlinear equations \eqref{NLEFX})
in Algorithm \ref{ALGCNMTR}. For the analysis of the global convergence and the
local suplinear convergence of Algorithm \ref{ALGCNMTR}, please refer to
references \cite{LXL2022,LX2021}.

\vskip 2mm

\begin{algorithm}
\renewcommand{\algorithmicrequire}{\textbf{Input:}}
\renewcommand{\algorithmicensure}{\textbf{Output:}}
\caption{Continuation Newton methods with the trust-region updating strategy
for finding a stationary point of the objective function  (CNMTr)}
\label{ALGCNMTR}
\begin{algorithmic}[1]
    \REQUIRE ~~ \\
    the gradient $g(x) = \nabla f(x)$ of the objective function
    $f: \; \Re^{n} \to \Re$, an initial point $x_0$ (optional),
    the tolerance $\epsilon$ (optional), the Hessian matrix $H(x)$
    of $f(x)$ (optional).
    \ENSURE ~~ \\
    The found stationary point $x^{\ast}$ of the objective function, the Boolean
    variable \textbf{success}.
    \STATE Set the default $(x_0)_{i} = 1 \; (i = 1, \, 2, \, \ldots, \, n)$ and
    $\epsilon = 10^{-6}$ when $x_0$ or $\epsilon$ is not provided.
    \STATE Initialize the parameters: $\eta_{a} = 10^{-6}, \; \eta_1 = 0.25, \; \eta_2 = 0.75, \;
    c_1 = 0.5, \; c_2 = 2, \; \overline{\Delta t} = 10^{-7}, \; \Delta t_{\text{init}} = 10^{-2}, \;
    \text{maxit} = 200$.
    \STATE Set $\Delta t_0 = \Delta t_{\text{init}}, \; \text{itc} = 0, \; k = 0$.
    \STATE Let the Boolean variable \textbf{success} be \textbf{false}.
    \STATE Let the Boolean variable \textbf{success\_{ap}} be \textbf{true}.
    \STATE Evaluate $g_{0} = \nabla f(x_0)$.
    \STATE Compute the residual $\text{Res}_0 = \|g_{0}\|_{\infty}$.
    \STATE Set $\rho_{-1}  = 0, \; s_{-1} = 0$.
    \WHILE{$\text{Res}_{k} > \epsilon$}
        \STATE Set $\text{itc} = \text{itc} + 1$.
        \IF{$\text{itc} > \text{maxit}$}
            \STATE break;
        \ENDIF
        \IF{\textbf{success\_{ap}} is \textbf{true}}
            \IF{$|1-\rho_{k-1}| > \eta_{1}$}
                \STATE Evaluate $H_k = H(x_{k})$.
            \ELSE
                 \STATE Set $H_{k} = H_{k-1}$.
            \ENDIF
            \STATE Solve  $H_{k} s_{k}^{N} = -g_{k}$ to obtain the Newton step
            $s_{k}^{N}$.
        \ENDIF
        \STATE Set $s_{k} = {\Delta t_{k}}/{(1+\Delta t_{k})} \, s_{k}^{N}, \;
        x_{k+1} = x_k + s_k$.
        \STATE Evaluate $g_{k+1} = \nabla f(x_{k+1})$.
        \STATE Compute the ratio $\rho_{k}$ from equation \eqref{RHOK}.
        \STATE Adjust the time step $\Delta t_{k+1}$ according to the
        trust-region updating strategy \eqref{TSK1}.
        \IF{$\rho_{k} \ge \eta_{a}$}
            \STATE Accept the trial point $x_{k+1}$.
            Compute the residual $\text{Res}_{k+1} = \|g_{k+1}\|_{\infty}$.
            \STATE Let the Boolean variable \textbf{success\_{ap}} be \textbf{true}.
        \ELSE
            \STATE Set $x_{k+1}  = x_{k}, \; g_{k+1} = g_{k}, \;
            s_{k+1}^{N} = s_{k}^{N}, \; \text{Res}_{k+1} = \text{Res}_{k}$.
            \STATE Let the Boolean variable \textbf{success\_{ap}} be \textbf{false}.
        \ENDIF
        \STATE Set $k \longleftarrow k+1$.
    \ENDWHILE
    \IF{$\text{Res}_{k} \le \epsilon$}
        \STATE Let the Boolean variable \textbf{success} be \textbf{true}.
    \ELSE
        \STATE Let the Boolean variable \textbf{success} be \textbf{false}.
    \ENDIF
    \STATE Set $x^{\ast} = x_{k}$.
    \STATE Output the found stationary point $x^{\ast}$ and the Boolean
    variable \textbf{success}.
\end{algorithmic}
\end{algorithm}

\vskip 2mm

In order to understand Algorithm \ref{ALGCNMTR} better, we give some explanations
of its operating mechanism and some key variables. In line 2, the parameters
$c_1, \; c_2, \; \eta_1, \; \eta_2, \; \overline{\Delta t}$ are used to adaptively
adjust the time step $\Delta t_{k}$ and they are defined by equation \eqref{TSK1}.
The ratio $\rho_{k}$ defined by \eqref{RHOK} represents the ratio of the actual
reduction $\|g(x_k)\| - \|g(x_k + s_k)\|$ to the predicted reduction
$\|g(x_k)\| - \|g(x_k) + H_{k}s_{k}\|$. The parameters $\eta_1, \; \eta_2$
($0 < \eta_1 < 1 \le \eta_2$) are used to define the range of the ratio
$\rho_{k}$ and indicate that the linear model $g(x_k) + H_{k}s_k$ approximates
$g(x_k + s_k)$ well or not. If $|1 - \rho_{k}| < \eta_{1}$, it indicates that
$g(x_{k}) + H_{k}s_{k}$ approximates $g(x_{k} + s_{k})$ well. Then, the time
step $\Delta t_{k+1}$ is enlarged and it equals $c_{2}\Delta t_{k}$. If
$|1 - \rho_{k}| > \eta_{2}$, it indicates that $g(x_{k}) + H_{k}s_{k}$
approximates $g(x_{k} + s_{k})$ bad. Then, the time step $\Delta t_{k+1}$ is
reduced and it equals $c_{1}\Delta t_{k}$. If
$\eta_{1} \le |1 - \rho_{k}| \le \eta_{2}$, it indicates that
$g(x_{k}) + H_{k}s_{k}$ approximates $g(x_{k} + s_{k})$ neither well nor bad.
Then, the time step $\Delta t_{k+1}$ is kept with $\Delta t_{k}$. The parameter
$0< \overline{\Delta t} < 1$ is a small number and used to prevent the time step
$\Delta t_{k}$ from becoming too small and even zero, when the gradient $g(x)$
is very nonlinear and $\Delta t_{k}$ is adaptively adjusted by the scheme
\eqref{TSK1}. $0 < \Delta t_{\text{init}} \le 1$ represents the initial time
step and its default value is $\Delta t_{\text{init}} = 10^{-2}$. The parameter
$\text{maxit}$ represents the maximum iterations.

\vskip 2mm

In line 4, the Boolean variable \textbf{success} is used to indicate that
the continuation Newton method (Algorithm \ref{ALGCNMTR}) finds a stationary
point of $f(x)$ successfully or not. When Algorithm \ref{ALGCNMTR} finds a
stationary point of $f(x)$ successfully, we let the Boolean variable
\textbf{success} be \textbf{true}. Otherwise, we let the Boolean variable
\textbf{success} be \textbf{false}. In line 5, the Boolean variable
\textbf{success\_{ap}} is used to indicate that the trial step $s_{k}$ is
accepted or not. When the trial step $s_{k}$ is accepted, we let the
Boolean variable \textbf{success\_{ap}} be \textbf{true}. Otherwise, we let the
Boolean variable \textbf{success\_{ap}} be \textbf{false}. In line 6, it
evaluates the gradient at the initial point $x_{0}$. In line 7, it computes the
infinite norm of the initial gradient. In lines 14-21, they give the procedures
to solve the Newton step $s_{k}^{N}$. In line 9, we use the infinite norm
$\text{Res}_{k} = \|g_{k}\|_{\infty} < \epsilon$ as the loop termination
condition, since it can provide more accurate assessment to the status of each
element $|g_{k}^{i}| < \epsilon \; (i = 1, \, 2, \ldots, \, n)$ than the 2-norm
$\|g_{k}\| < \epsilon$.

\vskip 2mm

In line 22, it computes the trial step $s_{k}$ and the trial point $x_{k+1}$.
In line 23, it evaluates the new gradient $g(x_{k+1})$. In lines 24, it computes
the ratio $\rho_{k}$ according to equation \eqref{RHOK}. In line 25, it adaptively
adjusts the time step $\Delta t_{k+1}$ according to the trust-region updating
strategy \eqref{TSK1}. In lines 26-32, they give the criterion whether the trial
step $s_{k}$ is accepted or not according to the ratio $\rho_{k} \ge \eta_{a}$
or not. In lines 35-39, they give a criterion how to assign the Boolean variable
\textbf{success}. When the final residual $\text{Res}_{k} = \|g_{k}\|_{\infty}$
is less than the given tolerance, we regard that Algorithm \ref{ALGCNMTR} finds
a stationary point of the objective function successfully and let the Boolean
variable \textbf{success} be \textbf{true}. Otherwise, we let the Boolean
variable \textbf{success} be \textbf{false}. In lines 40-41, they output the
found stationary point $x^{\ast} = x_{k}$ and the Boolean variable
\textbf{success}.

\vskip 2mm

\section{The deflation technique} \label{SECDT}

\vskip 2mm

By using Algorithm \ref{ALGCNMTR} (CNMTr) in Section \ref{SECCNM}, we can
find a stationary point $x_{1}^{\ast}$ of $f(x)$. Our strategy is that we
repeatedly use Algorithm \ref{ALGCNMTR} to find multiple stationary points of $f(x)$.
Then, we use these found stationary points as the initial evolutionary
seeds of the genetic algorithm in Section \ref{SECQGA}. If we directly use
Algorithm \ref{ALGCNMTR} with the multi-start method, a starting point could
easily lie in the basin of attraction of a previously found local minimum and
algorithm \ref{ALGCNMTR} may find the same local minimum even if those searches
are initiated from points that are far apart. In order to overcome this
disadvantage, we revise the deflation technique and combine it with Algorithm
\ref{ALGCNMTR} to find multiple stationary points of $f(x)$ as possible.

\vskip 2mm

Assume that $x_{k}^{\ast}$ is a zero point of the function $G_{k-1}(x)$.
Then, we construct another function $G_{k}(\cdot)$ via eliminating the zero point
$x_{k}^{\ast}$ of $G_{k-1}(x)$ as follows:
\begin{align}
  G_{k}(x) = \frac{1}{\|x-x_{k}^{\ast}\|} G_{k-1}(x), \; k =  1, \,
  2, \ldots, \label{FUNGK1}
\end{align}
where $G_{0}(x) = g(x) = \nabla f(x)$. Thus, from equation \eqref{FUNGK1}, we
obtain
\begin{align}
  G_{k}(x) = \frac{1}{\|x-x_{1}^{\ast}\| \cdots \|x-x_{k}^{\ast}\|} g(x),
  \label{FUNGKX}
\end{align}
where $x_{i}^{\ast} \, (i = 1, \, \ldots, \, k)$ are zeros of $g(x)$.
From equation \eqref{FUNGKX}, we know that the zero $x^{\ast}_{k+1}$ of
$G_{k}(x)$ is also the zero of $g(x)$. Namely, the zero set of $G_{k}(x)$
is a zero subset of $g(x)$.

\vskip 2mm

When $g(x_{i}^{\ast}) = 0$ and the Hessian matrix $H(x_{i}^{\ast})$ is
nonsingular, Brown and Gearhard \cite{BG1971} have proved
\begin{align}
    \lim_{j \to \infty} \inf \|G_{k}(x_{j})\| > 0 \nonumber
\end{align}
for any sequence $x_{j} \to x_{i}^{\ast}$, where $G_{k}(x)$ is defined by equation
\eqref{FUNGKX}. Actually, we can obtain the following estimation of the lower
bound of $G_{k}(x)$ when $x$ belongs to the neighborhood of $x_{i}^{\ast}$.

\begin{lemma} \label{LEMECP}
Assume that the gradient $g(x)$ of the objective function is continuously differentiable
and $x_{i}^{\ast}$ is a zero point of the gradient $g(\cdot)$. Furthermore, the Hessian
matrix $H(x_{i}^{\ast})$ satisfies
\begin{align}
    \|H(x_{i}^{\ast})y\| \ge c_{l} \|y\|,  \; \forall  \; y \in R^{n},
    \label{NONSIGJK}
\end{align}
where $c_{l}$ is a positive constant. Then, there exists a neighborhood
$B_{\delta_i}(x_{i}^{\ast}) = \{x: \; \|x - x_{i}^{\ast}\| \le \delta_i\}$
of $x_{i}^{\ast}$ such that
\begin{align}
    \|G_{k}(x)\| \ge \frac{c_l}{2} \frac{1}{\|x-x_{1}^{\ast}\| \cdots
    \|x-x_{i-1}^{\ast}\| \|x-x_{i+1}^{\ast}\| \cdots
    \|x-x_{k}^{\ast}\|}     \label{GKXGEPCNST}
\end{align}
holds for all $x \in B_{\delta_i}(x_{i}^{\ast}) \; (i = 1, \, 2, \, \ldots, \, k)$,
where $G_{k}(x)$ is defined by equation \eqref{FUNGKX}. That is to say,
 $x_{i}^{\ast} \; (i = 1, \, 2, \, \ldots, \, k)$ is not a zero of $G_{k}(x)$.
\end{lemma}
\proof Since $x_{i}^{\ast}$ is a zero of $g(x)$, according
to the first-order Taylor expansion, we have
\begin{align}
    g(x) & = g(x_{i}^{\ast}) + \int_{0}^{1}
    H(x_{i}^{\ast}+ t (x - x_{i}^{\ast})) (x - x_{i}^{\ast})dt \nonumber \\
    & = \int_{0}^{1} H(x_{i}^{\ast}+ t (x - x_{i}^{\ast})) (x - x_{i}^{\ast})dt.
   \label{FOTEXP}
\end{align}
According to the nonsingular assumption \eqref{NONSIGJK} of $H(x_{i}^{\ast})$
and the continuity of $H(\cdot)$, there exists a neighborhood
$B_{\delta_{i}}(x_{i}^{\ast}) = \{x: \|x-x_{i}^{\ast}\| \le \delta_{i}\}$
of $x_{i}^{\ast}$ such that
\begin{align}
    \|H(x) - H(x_{i}^{\ast})\| \le \frac{c_l}{2}   \label{JKCON}
\end{align}
holds for all  $ x \in B_{\delta_{i}}(x_{i}^{\ast})$. From equations
\eqref{FOTEXP}-\eqref{JKCON}, we have
\begin{align}
    & \|g(x) - H(x_{i}^{\ast})(x-x_{i}^{\ast})\| =
    \left\|\int_{0}^{1} (H(x_{i}^{\ast}+ t (x - x_{i}^{\ast})) - H(x_{i}^{\ast}))
    (x - x_{i}^{\ast})dt\right\| \nonumber \\
    \hskip 2mm & \le \int_{0}^{1}
    \|H(x_{i}^{\ast}+ t (x - x_{i}^{\ast})) - H(x_{i}^{\ast})\|
    \|x - x_{i}^{\ast}\| dt \le \frac{c_l}{2} \|x - x_{i}^{\ast}\|, \;
    \forall x \in B_{\delta_{i}}(x_{i}^{\ast}).
  \label{NSIGJKX}
\end{align}

\vskip 2mm

Furthermore, from the triangle inequality $\|x - y \| \ge \|x\| - \|y\|$ and
the assumption \eqref{NONSIGJK}, we have
\begin{align}
    \|g(x) - H(x_{i}^{\ast})(x-x_{i}^{\ast})\| \ge
    \|H(x_{i}^{\ast})(x-x_{i}^{\ast})\| - \|g(x)\| \ge
    c_{l} \|x-x_{i}^{\ast}\| - \|g(x)\|.   \label{TRINEQ}
\end{align}
Thus, from equations \eqref{NSIGJKX}-\eqref{TRINEQ}, we obtain
\begin{align}
    \|g(x)\| \ge \frac{c_l}{2} \|x-x_{i}^{\ast}\|, \;
    \forall x \in B_{\delta_{i}}(x_{i}^{\ast}).  \label{GKXGEPN}
\end{align}
By substituting inequality \eqref{GKXGEPN} into equation \eqref{FUNGKX}, we obtain
\begin{align}
    \|G_{k}(x)\| \ge \frac{c_l}{2} \frac{1}{\|x-x_{1}^{\ast}\| \cdots
    \|x-x_{i-1}^{\ast}\| \|x-x_{i+1}^{\ast}\| \cdots
    \|x-x_{k}^{\ast}\|} > 0,  \; \forall x \in B_{\delta_i}(x_{i}^{\ast}). \nonumber
\end{align}
That is to say, $x_{i}^{\ast} \; (i = 1, \, 2, \, \ldots, \, k)$ is not a zero
of $G_{k}(x)$. \qed

\vskip 2mm

In practice, we need to avoid the overflow or the underflow when we compute
$G_{k}(x)$, and we expect that the continuation Newton method (Algorithm
\ref{ALGCNMTR}) finds multiple stationary points of $f(x)$ as possible. According
to our numerical experiments, Algorithm \ref{ALGCNMTR} can find more stationary
points if we replace $\|x - x_{i}^{\ast}\| \; (i = 1, \, \ldots, \, k)$ with
$\|x - x_{i}^{\ast}\|_{1} \; (i = 1, \, \ldots, \, k)$ in equation
\eqref{FUNGKX}. We also notice that $\|x - x_{i}^{\ast}\|_{1} \approx
\|x_{i}^{\ast}\|_{1} \approx n$ when $x$ is close to zero, where $n$ is the
dimension of $x_{i}^{\ast}$. Thus, Algorithm \ref{ALGCNMTR} is apt to find
the spurious root of $G_{k}$ when the dimension $n$ is large. Therefore, in
order to improve the effect of the deflation technique, we modify
$G_{k}(x)$ defined by equation \eqref{FUNGK1} as
\begin{align}
    G_{k}(x) = \frac{\alpha_{k}}{\|x-x_{k}^{\ast}\|_{1}} G_{k-1}(x)
    = \left(\prod_{i = 1}^{k} \frac{\alpha_{i}}{\|x-x_{i}^{\ast}\|_{1}}\right) g(x),
    \label{SFUNGK1}
\end{align}
where $x_{i}^{\ast} \; (i = 1, \, 2, \, \ldots, \, k)$ are found zeros of
$g(x)$ and $\alpha_{i} \; (i = 1, \, \ldots, \, k)$ are defined by
\begin{align}
     \alpha_{i} =
         \begin{cases}
               n, \; \text{if} \; \|x_{i}^{\ast}\|_{1} \le 10^{-6}, \\
               \|x_{i}^{\ast}\|_{1}, \; \text{otherwise},
         \end{cases} \; i = 1, \, 2, \, \ldots, \, k.
         \label{ALPHAK}
\end{align}

\vskip 2mm

Since $G_{k}(x)$ has the special structure and its subdifferential exists except
for finite points $x_{1}^{\ast}, \, \ldots, \, x_{k}^{\ast}$, from equations
\eqref{SFUNGK1}-\eqref{ALPHAK}, we compute its subdifferential $\partial G_{k}(x)$
as follows:
\begin{align}
     & \partial G_{k} (x) =  \frac{\alpha_{1} \cdots \alpha_{k}}
     {\|x - x_{1}^{\ast}\|_{1} \cdots \|x-x_{k}^{\ast}\|_{1}}
     \frac{\partial g(x)}{\partial x} + \frac{\alpha_{1} \cdots \alpha_{k}}
     {\|x - x_{1}^{\ast}\|_{1} \cdots \|x-x_{k}^{\ast}\|_{1}}g(x)p_{k}(x)^{T}
     \nonumber  \\
     & \hskip 2mm
     = \frac{\alpha_{1} \cdots \alpha_{k}}
     {\|x - x_{1}^{\ast}\|_{1} \cdots \|x-x_{k}^{\ast}\|_{1}}
     \left(\frac{\partial g(x)}{\partial x} + g(x)p_{k}(x)^{T} \right), \label{GDGKF}
\end{align}
where $p_{k}(x)$ is defined by
\begin{align}
     p_{k}(x)^{T} = - \left(\frac{\text{sgn}\left(x - x_{1}^{\ast}\right)}
     {\|x - x_{1}^{\ast}\|_{1}} + \cdots + \frac{\text{sgn}\left(x - x_{k}^{\ast}\right)}
     {\|x - x_{k}^{\ast}\|_{1}}\right).  \label{VECFUNPX}
\end{align}
Here, the sign function $\text{sgn}(\cdot)$ is defined by
\begin{align}
     \text{sgn}(\beta) =
         \begin{cases}
             1, \; \text{if} \; \beta >  0, \\
             0, \; \text{if} \; \beta = 0, \\
             -1, \; \text{otherwise}. \label{SGNFUN}
         \end{cases}
\end{align}

\vskip 2mm

Now, by using the revised deflation technique \eqref{SFUNGK1} of $g(x)$ and the
continuation Newton method (Algorithm \ref{ALGCNMTR}), we can find a new stationary
point of $f(x)$ other than found stationary points. We describe their
detailed implementations in Algorithm \ref{ALGCNMDT}.

\vskip 2mm

\begin{algorithm}
\renewcommand{\algorithmicrequire}{\textbf{Input:}}
\renewcommand{\algorithmicensure}{\textbf{Output:}}
\caption{Continuation Newton methods with the deflation technique for finding
a new stationary point of the objective function (CNMDT)}
\label{ALGCNMDT}
\begin{algorithmic}[1]
    \REQUIRE ~~ \\
      the gradient $g(x) = \nabla f(x)$ of the objective function
      $f: \; \Re^{n} \to \Re$, $K$ known stationary points
      $x_{1}^{\ast}, \, \ldots, \, x_{K}^{\ast}$ of $f(x)$, an initial point $x_{0}$
      (optional), the tolerance $\epsilon$ (optional), the Hessian matrix $H(x)$
      of $f(x)$ (optional).
	\ENSURE ~~ \\
      $K+1$ found stationary points $x_{1}^{\ast}, \, \ldots, \, x_{K+1}^{\ast}$ including the newly
      found stationary point $x_{K+1}^{\ast}$ of $f(x)$, the Boolean variable \textbf{success\_dt}.
    \STATE Set the default $x_{0} = (1, \, \ldots, 1)^{T}$ and
    $\epsilon = 10^{-6}$ when $x_0$ or $\epsilon$ is not provided.
    \STATE Initialize the parameters: $\eta_{a} = 10^{-6}, \; \eta_1 = 0.25, \; \eta_2 = 0.75, \;
    c_1 = 0.5, \; c_2 = 2, \; \overline{\Delta t} = 10^{-7}, \; \Delta t_{\text{init}} = 10^{-2}, \;
    \text{maxit} = 200$.
    \STATE Let the Boolean variable \textbf{success\_dt} be \textbf{false}.
    \STATE Set $\Delta t_0 = \Delta t_{\text{init}}, \; \text{itc} = 0, \; k = 0$.
    \STATE Evaluate $g_{0} = \nabla f(x_0)$. Then, compute $F_{0} = G_{K}(x_0)$ from
    equations \eqref{SFUNGK1}-\eqref{ALPHAK}.
    \STATE Set $\rho_{-1}  = 0, \; s_{-1} = 0$.
    \STATE Let the Boolean variable \textbf{success\_{ap}} be \textbf{true}.
    \STATE Compute the residual $\text{Res}_{0} = \|F_{0}\|_{\infty}$.
    \WHILE{$\text{Res}_k > \epsilon$}
        \STATE $\text{itc} = \text{itc} + 1$.
        \IF{$\text{itc} > \text{maxit}$}
            \STATE break;
        \ENDIF
        \IF{\textbf{success\_{ap}} is true}
            \STATE Evaluate $H_k = H(x_{k})$. Then, compute the Jacobian
            $J_{k} = \partial G_{K}(x_k)$ from equations \eqref{GDGKF}-\eqref{SGNFUN}.
            \STATE Solve  $J_{k} s_{k}^{N} = -F_{k}$ to obtain the Newton step
            $s_{k}^{N}$.
        \ENDIF
        \STATE Set $s_{k} = {\Delta t_{k}}/{(1+\Delta t_{k})} \, s_{k}^{N}, \;
        x_{k+1} = x_k + s_k$.
        \STATE Evaluate $g_{k+1} = \nabla f(x_{k+1})$. Then, compute
        $F_{k+1} = G_{K}(x_{k+1})$ from equations \eqref{SFUNGK1}-\eqref{ALPHAK}.
        \STATE Compute the ratio $\rho_{k}$ from equation \eqref{RHOK}.
        \STATE Adjust the time step $\Delta t_{k+1}$ according to the trust-region
        updating strategy \eqref{TSK1}.
        \IF{$\rho_{k} \ge \eta_{a}$}
            \STATE Accept the trial point $x_{k+1}$.
            Compute the residual $\text{Res}_{k+1} = \|F_{k+1}\|_{\infty}$.
            \STATE Let the Boolean variable \textbf{success\_{ap}} be \textbf{true}.
        \ELSE
            \STATE Set $x_{k+1}  = x_{k}, \; F_{k+1} = F_{k}, \;
            s_{k+1}^{N} = s_{k}^{N}, \; \text{Res}_{k+1} = \text{Res}_{k}$.
            \STATE Let the Boolean variable \textbf{success\_{ap}} be \textbf{false}.
        \ENDIF
        \STATE Set $k \longleftarrow k+1$.
    \ENDWHILE
    \IF{$\text{Res}_{k} \le \epsilon$}
        \STATE Let the Boolean variable \textbf{success\_dt} be \textbf{true}.
        \STATE Save the newly found stationary point as $x_{K+1}^{\ast} = x_{k}$.
        \STATE Output $K+1$ found stationary points $x_{1}^{\ast}, \, \ldots, \, x_{K+1}^{\ast}$
        and the Boolean variable \textbf{success\_dt}.
    \ELSE
        \STATE Let the Boolean variable \textbf{success\_dt} be \textbf{false}.
        \STATE Output $K$ known stationary points $x_{1}^{\ast}, \, \ldots, \, x_{K}^{\ast}$
        and the Boolean variable \textbf{success\_dt}.
    \ENDIF
\end{algorithmic}
\end{algorithm}

In order to understand Algorithm \ref{ALGCNMDT} better, we give some explanations
of its operating mechanism and its key variables. $K$ is the number of found
stationary points of $f(x)$. In lines 1-2, they give the default parameters used
by the continuation Newton method to find a new stationary point of $f(x)$ and
they are explained in Algorithm \ref{ALGCNMTR}. In line 3, the Boolean variable
\textbf{success\_dt} is used to indicate whether the continuation Newton method
finds a new stationary point successfully or not. When the continuation Newton
finds a new stationary point successfully, we let the Boolean variable
\textbf{success\_dt} be \textbf{true}. Otherwise, we let Boolean variable
\textbf{success\_dt} be \textbf{false}.

\vskip 2mm

In lines 4-30, they give the pseudo codes of the continuation Newton method
(CNMTr) to find a new stationary point of $f(x)$. They are equivalent to
Algorithm \ref{ALGCNMTR} if we replace $g(x)$ and $H(x)$ with $G_{K}(x)$ and
$\partial G_{K}(x)$ in Algorithm \ref{ALGCNMTR}, respectively. For their
explanations, please refer to the related explanations in Algorithm
\ref{ALGCNMTR}. In lines 31-38, they output different values according to
whether a new stationary point is found successfully or not. When
$\text{Res}_{k} \le  \epsilon$, it indicates that the continuation Newton method
finds a new stationary point successfully and we let the Boolean variable
\textbf{success\_dt} be \textbf{true}. Then, this newly found stationary point
is saved as $x_{K+1}^{\ast}$. They output $K+1$ found stationary points
$x_{1}^{\ast}, \,\ldots, \, x_{K+1}^{\ast}$ and the Boolean variable
\textbf{success\_dt}. Otherwise, We let the Boolean variable
\textbf{success\_dt} be \textbf{false}. Then, they output $K$ known stationary
points $x_{1}^{\ast}, \, \ldots, \, x_{K}^{\ast}$ and the Boolean variable
\textbf{success\_dt}.

\vskip 2mm

\begin{remark} From equation \eqref{SFUNGK1}, we know that $\nabla f(x_{l})$
may converge to a positive number when
$\lim_{l \to \infty} \|x_{l}-x_{i}^{\ast}\|_{1} = \infty$, where $x_{i}^{\ast}$
is the stationary point of $f(x)$. One of the reasons is that the Newton method
fails to converge to the certain stationary point of $f(x)$ when the initial
point $x_{0}$ belongs to a special region \cite{Abbott1977}. Thus, the
continuation Newton method with the deflation technique may not find all
stationary points of $f(x)$ from the same initial point $x_{0}$.
\end{remark}

\vskip 2mm

In order to find multiple stationary points of $f(x)$ as possible, we use the
continuation Newton method with the deflation technique (i.e. Algorithm
\ref{ALGCNMDT}) to find stationary points from several initial points that
are far apart. We define the $\frac{n}{2}$-dimensional vector
$e = (1, \, \ldots, \, 1)^{T}$ and select the following six points
$x_{i}^{\text{init}}\; (i = 1, \, \ldots, \, 6)$ as the default initial points:
\begin{align}
    & x_{1}^{\text{init}} = \dbinom{e}{e}, \; x_{2}^{\text{init}}= - \dbinom{e}{e},
    \; x_{3}^{\text{init}} = \dbinom{e}{-e},
     \;  x_{4}^{\text{init}} = \dbinom{-e}{e}, \nonumber     \\
    & x_{5}^{\text{init}}(i) = i, \; i = 1, \, 2, \, \ldots, \, n, \;
    x_{6}^{\text{init}}(i) = (n+1) - i, \; i = 1, \, 2, \, \ldots, \, n,  \label{INITALG}
\end{align}
where $x_{j}^{\text{init}}(i)$ represents the $i$-th element of the $n$-dimensional
vector $x_{j}^{\text{init}}$. Thus, we repeatedly use Algorithm \ref{ALGCNMDT} from
multi-start points to find multiple stationary points of $f(x)$.
We describe their detailed implementations in Algorithm \ref{ALGCNMDTM}.

\vskip 2mm

\begin{algorithm}
\renewcommand{\algorithmicrequire}{\textbf{Input:}}
\renewcommand{\algorithmicensure}{\textbf{Output:}}
\caption{Continuation Newton methods with the deflation technique and
 multi-start points for finding multiple stationary points (CNMDTM)}
  \label{ALGCNMDTM}
\begin{algorithmic}[1]
    \REQUIRE ~~ \\
      the gradient $g(x) = \nabla f(x)$ of the objective function
      $f: \; \Re^{n} \to \Re$, the tolerance $\epsilon$ (optional), $L$ initial
      points $x_{1}^{\text{init}}, \, \ldots, \, x_{L}^{\text{init}}$ (optional),
      the Hessian matrix $H(x)$ of $f(x)$ (optional).
	\ENSURE ~~ \\
      All found stationary points $x_{1}^{\ast}, \, \ldots, \, x_{\textbf{nsp}}^{\ast}$
      of $f(x)$, and its global optimal approximation point $x_{ap}^{\ast}$.
    \STATE Set the default tolerance $\epsilon = 10^{-6}$.
    \STATE Set the default initial points $x_{i}^{\text{init}} \;
    (i = 1, \, \ldots, \, L)$ given by equation \eqref{INITALG}, where $L = 6$.
    \STATE Let the Boolean variable \textbf{success} be \textbf{false}.
    \STATE Set $\textbf{nsp} = 0$ (\textbf{nsp} represents the number of found
    stationary points of $f(x)$).
    \FOR{$i = 1, \, \ldots, \, L$}
        \STATE Select the $i$-th initial point as $x_{0} = x_{i}^{\text{init}}$
        and call Algorithm \ref{ALGCNMTR} to find a stationary point of $f(x)$.
        \IF{a stationary point of $f(x)$ is found by Algorithm \ref{ALGCNMTR} successfully}
            \STATE Set $\textbf{nsp} = \textbf{nsp} + 1$.
            \STATE Let the Boolean variable \textbf{success} be \textbf{true}.
            \STATE Save this found stationary point of $f(x)$ as $x_{1}^{\ast}$.
            \STATE break;
        \ELSE
            \STATE Let the Boolean variable \textbf{success} be \textbf{false}.
        \ENDIF
    \ENDFOR
    \IF{the Boolean variable \textbf{success} is \textbf{true}}
        \STATE Set $i = 1$ ($i$ represents the order of given initial points).
        \WHILE{$i \le L$}
            \STATE Select the $i$-th initial point as $x_{0} = x_{i}^{\text{init}}$
            and call Algorithm \ref{ALGCNMDT} to find a new stationary point.
            \IF{a new stationary point of $f(x)$ is found successfully}
                \STATE Set $\textbf{nsp} = \textbf{nsp} + 1$.
                \STATE Save this newly found stationary point as $x_{\textbf{nsp}}^{\ast}$.
                \IF{this newly found stationary point $x_{\textbf{nsp}}^{\ast}$
                almost equals the initial point $x_{i}^{\text{init}}$}
                    \STATE Set $i = i+1$.
                \ENDIF
            \ELSE
                \STATE Set $i = i+1$.
            \ENDIF
        \ENDWHILE
    \ENDIF
    \IF{$\textbf{nsp} > 0$}
        \STATE Set $f_{ap}^{\ast} = f(x_{1}^{\ast})$ and $x_{ap}^{\ast} = x_{1}^{\ast}$.
        \FOR{$k = 1, \, \ldots, \, \textbf{nsp}$}
            \STATE Set $f_{k}^{\ast} = f(x_{k}^{\ast})$.
            \IF{$f_{k}^{\ast} < f_{ap}^{\ast}$}
                \STATE Set $x_{ap}^{\ast} = x_{k}^{\ast}$ and $f_{ap}^{\ast} = f_{k}^{\ast}$.
            \ENDIF
       \ENDFOR
       \STATE Let the Boolean variable \textbf{success} be \textbf{true}.
       \STATE Output all found stationary points
       $x_{1}^{\ast}, \, \ldots, \, x_{\textbf{nsp}}^{\ast}$ of $f(x)$, its global optimal
       approximation point $x_{ap}^{\ast}$, its optimal approximation value $f_{ap}^{\ast}$,
       and the Boolean variable \textbf{success}.
    \ELSE
       \STATE Set $x_{ap}^{\ast} = x_{0}$ and $f_{ap}^{\ast} = f(x_0)$.
       \STATE  Let the Boolean variable \textbf{success} be \textbf{false}.
       \STATE Output $x_{ap}^{\ast}, \; f_{ap}^{\ast}$ and the Boolean variable
       \textbf{success}.
    \ENDIF
\end{algorithmic}
\end{algorithm}

\vskip 2mm

In order to understand Algorithm \ref{ALGCNMDTM} better, we give some explanations
of its operating mechanism and its key variables. In line 1,
$x_{1}^{\text{init}}, \, \ldots, \, x_{L}^{\text{init}}$ are $L$ default initial
points defined by equation \eqref{INITALG}, where $L = 6$. In line 3, the
Boolean variable \textbf{success} represents whether a stationary point of $f(x)$
is found by CNMTr (Algorithm \ref{ALGCNMTR}) successfully or not. When Algorithm
\ref{ALGCNMTR} (CNMTr) finds a stationary point successfully, we let the Boolean
variable \textbf{success} be \textbf{true}. Otherwise, we let the Boolean variable
\textbf{success} be \textbf{false}. In line 4, \textbf{nsp} represents the number
of found stationary points and its default value equals 0.

\vskip 2mm

In lines 5-15, they repeatedly call Algorithm \ref{ALGCNMTR} (CNMTr) to find a
stationary point of $f(x)$ from $L$ default initial points
$x_{1}^{\text{init}}, \, \ldots, \, x_{L}^{\text{init}}$. Once Algorithm
\ref{ALGCNMTR} finds a stationary point, we let the Boolean variable
\textbf{success} be \textbf{true}, set $\textbf{nsp} = \textbf{nsp} + 1$, save
this found stationary point as $x_{1}^{\ast}$,  and exit the loop. Otherwise,
we let the Boolean variable \textbf{success} be \textbf{false}.

\vskip 2mm

In lines 16-30, if Algorithm \ref{ALGCNMTR} finds a stationary point of $f(x)$
successfully, they will repeatedly call Algorithm \ref{ALGCNMDT} (CNMDT) to
find multiple stationary points of $f(x)$ from $L$ default initial points
$x_{1}^{\text{init}}, \, \ldots, \, x_{L}^{\text{init}}$.
In line 17, the variable $i$ represents the order of given initial points.
In line 18, it means that we execute lines 19-28 if the order variable $i$
is less than the number of given initial points (i.e. $L$). In line 19, it
selects the $i$-th initial point $x_{i}^{\text{init}}$ and calls Algorithm
\ref{ALGCNMDT} (CNMDT) to find a new stationary point of $f(x)$.
In lines 20-28, they execute the different procedures according to whether
a new stationary point is found successfully by Algorithm \ref{ALGCNMDT}
(CNMDT) or not. If a new stationary point of $f(x)$ is found Algorithm
\ref{ALGCNMDT} (CNMDT) successfully, we let $\textbf{nsp}$ (the number of found
stationary points) increase by 1 and save this newly found stationary point.
Otherwise, we let the order variable $i$ increase by 1 and select the next
initial point. In lines 22-23, they mean that we need to select the next
initial point if the newly found stationary point almost equals the $i$-th
initial point $x_{i}^{\text{init}}$.

\vskip 2mm

In lines 31-45, if \textbf{nsp} (the number of found stationary points)
is greater than 0, we let the Boolean variable \textbf{success} be
\textbf{true} and output all found stationary points $x_{1}^{\ast}, \, \ldots,
\, x_{\text{nsp}}^{\ast}$, the optimal approximation point
$x_{ap}^{\ast} = \arg \min \{f(x_{1}^{\ast}), \, \ldots,
\, f(x_{\textbf{nsp}}^{\ast})\}$, and the Boolean variable \textbf{success}.
Otherwise, we let Boolean variable \textbf{success} be \textbf{false} and
output the optimal approximation point $x_{ap}^{\ast} = x_{0}$, and the Boolean
variable \textbf{success}.

\vskip 2mm

\section{The quasi-genetic evolution} \label{SECQGA}

\vskip 2mm

By using the continuation Newton method with the deflation technique and
multi-start points (Algorithm \ref{ALGCNMDTM}) in Section \ref{SECDT}, we find
the most of stationary points $x_{1}^{\ast}, \, \ldots, \, x_{\textbf{nsp}}^{\ast}$
and obtain the suboptimal value $f(x_{ap}^{\ast})$. Algorithm
\ref{ALGCNMDTM} is essentially a local search method, and it can not ensure that
$f(x_{ap}^{\ast})$ is the global minimum of $f(x)$ except that all stationary
points of $f(x)$ have been found by Algorithm \ref{ALGCNMDTM} (that would imply
$f(x_{ap}^{\ast})$ is a global minimum). Therefore, in order
to find the global minimum at all possible, we use the idea of memetic algorithms
\cite{Moscato1989,SGKZ2014} to approach the global minimum further. Namely, we use
the property that evolutionary algorithms escape from the local minima. And we
revise the heuristic crossover of the genetic algorithm such that it evolves from
the found stationary points $x_{1}^{\ast}, \, \ldots, \, x_{\textbf{nsp}}^{\ast}$
of $f(x)$. After it evolves several generations such as twenty generations, we
obtain an evolutionary suboptimal solution $x_{\text{ag}}^{\ast}$ of $f(x)$. Then,
we use it as the initial point of the continuation Newton method (i.e. CNMTr,
Algorithm \ref{ALGCNMTR}) to refine it and obtain a better approximation solution
$x_{\min}^{\ast}$ of the global minimum point.

\vskip 2mm

In order to ensure the effect of the evolutionary algorithm, the number of
population (we denote it as $L$) can not be too small. In general, $L$ is
greater than 20. Since the number of found stationary points may be less than $L$,
we supplement $L$ points as the candidate seeds of the evolutionary algorithm.

\vskip 2mm

We define the $\frac{n}{2}$-dimensional vector $e = (1, \, \ldots, \, 1)^{T}$. Then,
we select the following five basic seeds:
\begin{align}
   x_{0}^{bs} = 0, \quad x_{1}^{bs} = \dbinom{e}{e},  \quad
   x_{2}^{bs} = \dbinom{e}{-e},  \quad  x_{3}^{bs} = \dbinom{-e}{e}, \quad
   x_{4}^{bs} = -\dbinom{e}{e}. \label{BSINIT}
\end{align}
We denote the set $S_{0}^{bs}$ as
\begin{align}
     S_{0}^{bs} = \left\{x_{1}^{bs}, \;
    x_{2}^{bs}, \; x_{3}^{bs}, \; x_{4}^{bs}\right\}. \label{BSEEDSET}
\end{align}
From equations \eqref{BSINIT}-\eqref{BSEEDSET}, we supplement $L$ candidate seeds such as
\begin{align}
     S_{0}^{\text{seed}} = \left\{0, \, 10^{-1} \times S_{0}^{bs}, \, S_{0}^{bs}, \,
     10 \times S_{0}^{bs}, \, 10^{2} \times S_{0}^{bs}, \, 10^{3} \times S_{0}^{bs},
     \, \ldots \right\}. \label{IDCSEED}
\end{align}
In practice, we select the first $L$ minimum points of the objective
function from the set
\begin{align}
    \bar{S}_{0} = \left\{x_{1}^{\ast}, \, \ldots, x_{\textbf{nsp}}^{\ast}, \,
    S_{0}^{\text{seed}}\right\} \label{INICSEED}
\end{align}
and denote these $L$ points as $x_{i}^{0} \; (i = 1, \, \ldots, \, L)$, where
the set $S_{0}^{\text{seed}}$ is defined by equation \eqref{IDCSEED}. We denote
\begin{align}
    S_{0} = \left\{x_{1}^{0}, \, \ldots, \, x_{L}^{0}\right\} \label{SEEDSET}
\end{align}
and adopt them as the initial seeds of the quasi-genetic algorithm.

\vskip 2mm

Secondly, we select any two elements $x_{i}^{0}$ and $x_{j}^{0}$ from the
seed set $S_{0}$, and use the convex crossover to generate the next
offspring. Namely, we set
\begin{align}
    \bar{x}_{k}^{1} = \frac{1}{2}\left(x_{i}^{0} + x_{j}^{0}\right),
    \; i = 1, \, \ldots, \, L, \; j = i, \, i+1, \, \ldots, \, L, \;
    \; k = 1, \, 2, \ldots, \, \frac{1}{2}L(L-1).     \label{COVERGA}
\end{align}
Similarly, we select the first $L$ minimum points of $f(x)$ from the set
\begin{align}
    \bar{S}_{1} = S_{0} \bigcup \left\{\bar{x}_{1}^{1}, \, \ldots, \,
    \bar{x}_{L(L-1)/2}^{1}\right\} \label{MEGESETS1}
\end{align} and denote these $L$ points as
$x_{i}^{1} \, (i = 1, \, \ldots, \, L)$. We set
\begin{align}
    S_{1} = \left\{x_{1}^{1}, \, \ldots, \, x_{L}^{1} \right\} \label{SEEDS1}
\end{align} and regard them as the dominant population.

\vskip 2mm

Repeatedly, after the $l$-th evolution, we generate the offspring
$\left\{\bar{x}_{1}^{l+1}, \, \ldots, \, \bar{x}_{L(L-1)/2}^{l+1}\right\}$
and select the first $L$ minimum points of $f(x)$ from the set
\begin{align}
    \bar{S}_{l+1} = S_{l} \bigcup \left\{\bar{x}_{1}^{l+1}, \, \ldots, \,
    \bar{x}_{L(L-1)/2}^{l+l}\right\}, \; l = 0, \, 1, \ldots. \label{LGESEED}
\end{align}
We denote these $L$ points as $x_{i}^{l} \, (i = 1, \, \ldots, \, L)$, and set
\begin{align}
    S_{l+1} = \left\{x_{1}^{l+1}, \, \ldots, \, x_{L}^{l+1} \right\}, \;
    l = 0, \, 1, \, \ldots.  \label{SEEDSL}
\end{align}
After it evolves $N$ generations, we stop the evolution and select the minimum
point of $f(x)$ from the set $S_{N}$. We denote this minimum point as
$x_{ag}^{\ast}$.

\vskip 2mm

Finally, in order to improve the convergence of the quasi-genetic evolution,
we call Algorithm \ref{ALGCNMTR} (CNMTr) from $x_{ag}^{\ast}$ to find the
stationary point of $f(x)$. We denote this found stationary point as
$x_{cn}^{\ast}$. Then, we select the minimum point $x_{\min}^{\ast}$ of $f(x)$
between $x_{cn}^{\ast}$ and $x_{ag}^{\ast}$, i.e.
\begin{align}
   x_{\min}^{\ast} = \arg \min \left\{f(x_{cn}^{\ast}), \; f(x_{ag}^{\ast}) \right\},
   \nonumber
\end{align}
and output $x_{\min}^{\ast}$ as the global optimal approximation point of $f(x)$.
According to the above discussions, we give the detailed descriptions of the
quasi-genetic algorithm in Algorithm \ref{ALGQGE}.

\vskip 2mm

\begin{algorithm}
\renewcommand{\algorithmicrequire}{\textbf{Input:}}
\renewcommand{\algorithmicensure}{\textbf{Output:}}
\caption{The quasi-genetic evolution for global optimization problems
(QGE)}
\label{ALGQGE}
\begin{algorithmic}[1]
     \REQUIRE ~~ \\
      the gradient $g(x) = \nabla f(x)$ of $f(x)$, the known stationary points
      $x_{1}^{\ast}, \, \ldots, \, x_{\textbf{nsp}}^{\ast}$ of $f(x)$ and $L$ initial points
      $x_{1}^{\text{init}}, \, \ldots, \, x_{L}^{\text{init}}$ (optional), the
      tolerance $\epsilon$ (optional), the Hessian matrix $H(x)$ of $f(x)$ (optional).
	  \ENSURE ~~ \\
      the global optimal approximation point $x_{\min}$ of $f(x)$.  \\
    \STATE Set the default tolerance $\epsilon = 10^{-6}$.
    \STATE Set $L = 21$ and select the default initial points $x_{i}^{\text{init}} \;
    (i = 1, \, \ldots, \, L)$ from $S_{0}^{\text{seed}}$ defined by
    equation \eqref{IDCSEED}.
    \STATE Set $l = 0$ and $N = 20$ ($N$ is the maximum number of
    evolutionary generations).
     \STATE Select the first $L$ minimum points of $f(x)$ from the set
     $\bar{S}_{0}$ defined by equation \eqref{INICSEED}, and denote these $L$
     points as $x_{i}^{0} \; (i = 1, \, \ldots, \, L)$.
     \STATE Set $S_{0} = \left\{x_{1}^{0}, \, \ldots, \, x_{L}^{0} \right\}$.
     \WHILE{($l < N$)}
        \STATE Set $k = 1$.
        \FOR {$i = 1, \, \ldots, \, L$}
            \FOR {$j = (i + 1), \, \ldots, \, $}
                \STATE Set $\bar{x}_{k}^{l+1} =
                \frac{1}{2} \left(x_{i}^{l}+x_{j}^{l}\right)$.
                \STATE Set $k = k+1$.
            \ENDFOR
        \ENDFOR
        \STATE Select the first $L$ minimum points of $f(x)$ from the set
        $\bar{S}_{l+1}$ defined by equation \eqref{LGESEED}, and denote these
        $L$ points as $x_{i}^{l+1} \, (i = 1, \, \ldots, \, L)$.
        \STATE Set $S_{l+1} = \left\{x_{1}^{l+1}, \, \ldots, \,
        x_{L}^{l+1}\right\}$.
        \STATE Set $l \longleftarrow l +1$.
     \ENDWHILE
     \STATE Set $f_{ag}^{\ast} = f\left(x_{1}^{N}\right)$
       and $x_{ag}^{\ast} = x_{1}^{N}$.
     \STATE Use $x_{ag}^{\ast}$ as the initial point and call Algorithm
     \ref{ALGCNMTR} to find a stationary point $x_{cn}^{\ast}$ of $f(x)$.
     \STATE Set $x_{\min}^{\ast} =
     \arg \min \left\{f(x_{cn}^{\ast}), \; f(x_{ag}^{\ast}) \right\}$.
     \STATE Output $x_{\min}^{\ast}$ as the global optimal approximation point.
\end{algorithmic}
\end{algorithm}

\vskip 2mm

In order to understand Algorithm \ref{ALGQGE} better, we describes how it works
step by step and give some explanations of the key variables. \textbf{nsp}
represents the number of the known stationary points of $f(x)$. In line 2, $L$
represents the number of population and its default value equals $21$. It
supplements $L$ candidate seeds $x_{1}^{\text{init}}, \, \ldots, \,
x_{L}^{\text{init}}$ from the set $S_{0}^{\text{seed}}$ defined by equation
\eqref{IDCSEED}. In line 3, $N$ represents the maximum number of evolutionary
generations, and its default value equals $20$. In line 4, it selects the first
$L$ minimum points of $f(x)$ from the set $\bar{S}_{0}$ defined by equation
\eqref{INICSEED} as the initial seeds of the evolutionary algorithm, and denotes
the first $L$ minimum points as $x_{1}^{0}, \, \ldots, \, x_{L}^{0}$. In line 5,
we denote the set $S_{0}$ of the initial seeds as
$S_{0} = \{x_{1}^{0}, \, \ldots, \, \, x_{L}^{0}\}$.

\vskip 2mm

In lines 6-17, they give the crossover operation to generate the next offspring,
and select the first $L$ dominant offspring as the evolutionary population. After
it evolves $N$ generations, the evolution will stop. In line 18, it selects the
minimum point $x_{ag}^{\ast}$ from the final evolutionary offspring
$\left\{x_{1}^{N}, \, \ldots, \, x_{L}^{N}\right\}$. In line 19, it uses
$x_{ag}^{\ast}$ as the initial point of Algorithm \ref{ALGCNMTR} (CNMTr) and call
Algorithm \ref{ALGCNMTR} to find a stationary point $x_{cn}^{\ast}$ of $f(x)$.
In lines 20-21, they select the minimum from
$\{f(x_{ag}^{\ast}), \, f(x_{cn}^{\ast})\}$, and denote this minimum as
$f(x_{\min}^{\ast})$. Finally, it outputs $x_{\min}^{\ast}$ as the global optimal
approximation solution of $f(x)$.

\section{The implementation of CNMGE}

\begin{figure}[htbp]
\centering
\includegraphics[width=0.95\textwidth]{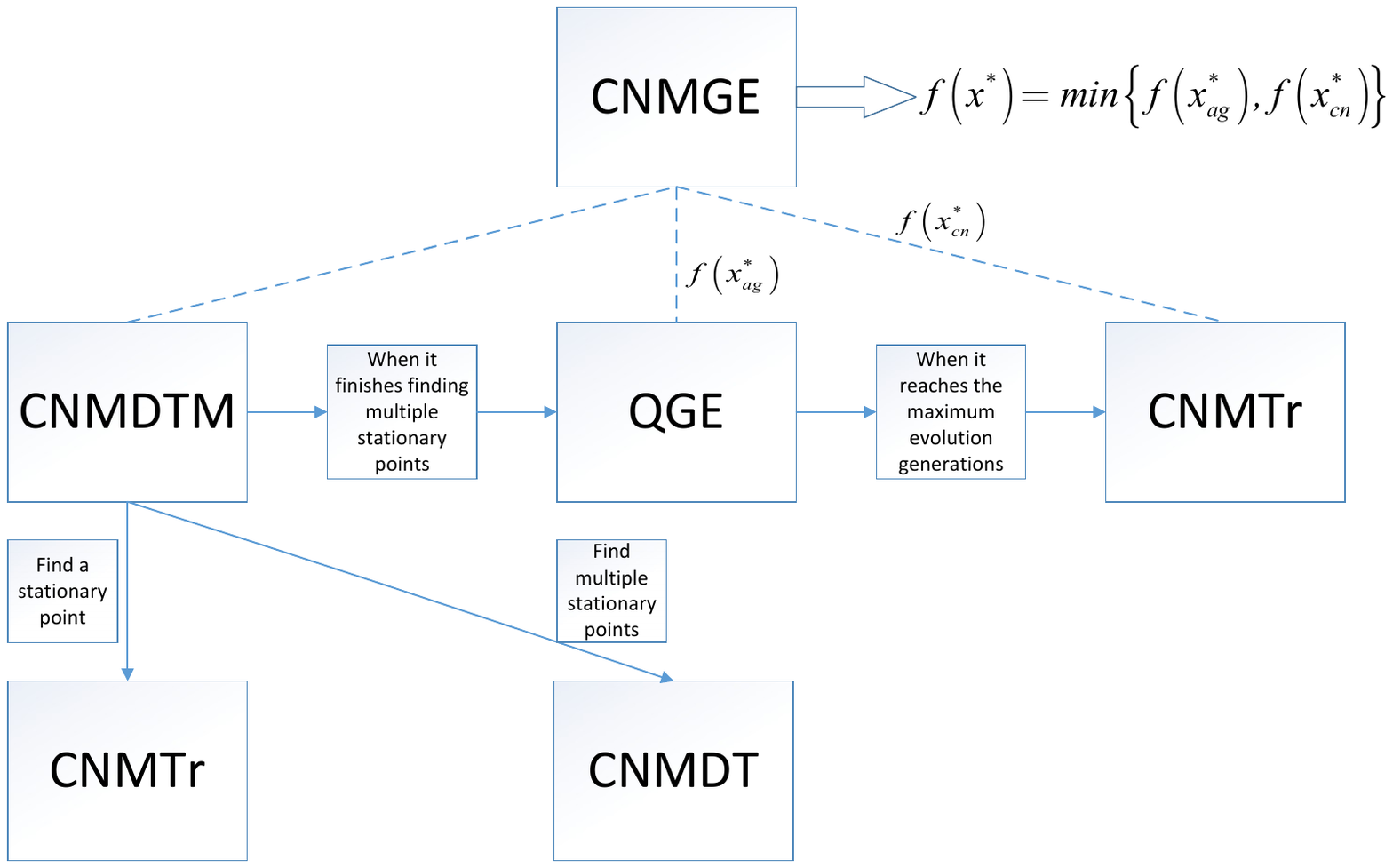}
\caption{The implementation diagram of CNMGE.}
\label{FIGCNMGE}
\end{figure}

In this section, according to the discussions of previous sections, we give
the implementation descriptions of the continuation Newton method with the
deflation technique and the quasi-genetic evolution for global optimization
problems. We denote this method as CNMGE. Its implementation diagram is
illustrated by Figure \ref{FIGCNMGE}. Firstly, CNMGE calls the subroutine CNMDTM
(Algorithm \ref{ALGCNMDTM}) to find multiple stationary points of $f(x)$
from multi-start points. Then, it calls QGE (Algorithm
\ref{ALGQGE}) and uses those found stationary points as the initial seeds
of QGE to approach the global optimal approximation $x_{ag}^{\ast}$ of
$f(x)$. Finally, it calls CNMTr (Algorithm \ref{ALGCNMTR}) to refine
$x_{ag}^{\ast}$ and obtain its refined point $x_{cn}^{\ast}$. It outputs
$x^{\ast} = \arg \min\{f(x_{ag}^{\ast}), \; f(x_{cn}^{\ast})\}$ as the global
optimal approximation point of $f(x)$. From Figure \ref{FIGCNMGE} and Algorithm
\ref{ALGCNMDTM}, we know that the subroutine CNMDTM firstly calls the subroutine
CNMTr (Algorithm \ref{ALGCNMTR}) to find a stationary point. Then, it repeatedly
calls the subroutine CNMDT (Algorithm \ref{ALGCNMDT}) to find a new stationary
point of $f(x)$ from multi-start points, and saves all found stationary points
$x_{1}^{\ast}, \, \ldots, \, x_{\textbf{nsp}}^{\ast}$ as the input of the subroutine
QGE.

\vskip 2mm

In order to retain the usability of derivative-free methods, we use the automatic
differentiation technique \cite{BPRS2017,GW2008,Neidinger2010,WV2013,WBB2014}
with the reverse mode to compute the gradient $\nabla f(x)$ of $f(x)$ in Algorithm
\ref{ALGCNMTR} (CNMTr) and Algorithm \ref{ALGCNMDT} (CNMDT). In general, the
time that it takes to calculate the gradient of the objective function $f(x)$ by
automatic differentiation with the reverse mode is about a constant (less than
four) multiple of the function cost \cite{GW2008}. We implement it via calling
the built-in subroutine prob2struct.m of MATLAB 2021b. At the end of this
section, we give an example to show how it works.

\vskip 2mm

In Algorithm \ref{ALGCNMTR} and Algorithm \ref{ALGCNMDT}, in order to find a
stationary point of $f(x)$, it requires the Hessian matrix $H(x)$ of $f(x)$.
In practice, we replace the Hessian matrix $H(x)$ with its finite difference
approximation as follows:
\begin{align}
     H(x_{k}) \approx
    \left[\frac{g(x_{k} + \epsilon e_{1}) - g(x_{k})}{\epsilon}, \,
    \ldots, \, \frac{g(x_{k} + \epsilon e_{n}) - g(x_{k})}{\epsilon}\right],
    \label{NUMHESS}
\end{align}
where $e_{i}$ represents the unit vector whose elements equal zeros except for
the $i$-th element which equals 1, and $\epsilon = 2\times 10^{-8}$.

\vskip 2mm

It is worthwhile to discuss the selection of parameters in Algorithm
\ref{ALGCNMTR} and Algorithm \ref{ALGCNMDT}. In practice, we select
$\eta_{1} = 1/4, \; \eta_{2} = 3/4, \; c_{1} = 1/2, \; c_{2} = 2$ as the default
parameters. According to our numerical experiments, these default parameters work
well. Since the continuation Newton flow defined by equation \eqref{DAEFLOW} varies
dramatically in the transient phase, in order to follow its trajectory accurately,
we choose the small initial time step $\Delta t_{\text{init}}$ such as
$\Delta t_{\text{init}} = 10^{-2}$. This selection strategy of the initial time
step is different to that of the line search method. The initial step length of
the line search method is tried from $\alpha_{0} = 1$ such that it mimics the
Newton method and achieves the fast convergence rate near the stationary
point \cite{NW1999,SY2006}.

\vskip 2mm

We write a software package to implement CNMGE with the MATLAB language and it
can be executed in the MATLAB 2020b environment or after this version. For ease
of use, we give an example to illustrate how to use it.

\vskip 2mm

\noindent \textbf{Rosenbrock function} \cite{Rosenbrock1960}
    \begin{align}
        f(x) = 100(x_{2} - x_{1}^{2})^{2} + (x_{1} - 1)^{2}. \label{ROSENFUN}
    \end{align}
It is not difficult to know that the global minimum of the Rosenbrock
function \eqref{ROSENFUN} is located at $x^{\ast} = (1, \; 1)$ and
$f(x^{\ast}) = 0$. Its user guide is described in \textbf{Example 5.1} and
we give some explanations as follows.

\vskip 2mm

\begin{algorithm}
   \textbf{Example 5.1} The user guide of CNMGE \\
   \hrule
   \begin{algorithmic}[1]
       \STATE  $n = 2$;
       \STATE $x_0 = 2*\text{ones}(n, 1)$;
       \STATE \% Define the objective function.
       \STATE  $x$ = \text{optimvar}(`$x$', $n$, 1);
       \STATE $\text{objfun} = 100*(x(2)^\wedge{2}- x(1))^\wedge{2}
        + (x(1) - 1)^\wedge{2}$;
       \STATE  $lb = - \inf$;
       \STATE $ub = \inf$;
       \STATE $\text{unitdisk} = x(1)^\wedge{2} <= 1$;
       \STATE prob = optimproblem(``Objective'', objfun, ``Constraints'', unitdisk);
       \STATE problem = prob2struct(prob,``ObjectiveDerivative'',``auto-reverse'');
       \STATE [$x$\_opt, $f$\_opt, CPU\_time]
         = \text{CNMGE}(\text{problem.objective}, $x_0$, $ub$, $lb$);
       \STATE \% Output the global minimum value computed by CNMGE.
       \STATE fprintf(``The global minimum value computed by
        CNMGE is \%12.8f $\backslash n$'', $f$\_opt);
       \STATE \% Output the computational time of CNMGE.
       \STATE fprintf(``The computational time of CNMGE is \%8.4f$\backslash n$'', CPU\_time);
   \end{algorithmic}
\end{algorithm}

In line 4, the subroutine \textbf{optimvar} defines an $n$-dimensional vector. In line 5,
it defines an objective function such as $f(x) = 100(x_{2} - x_{1}^{2})^{2} + (x_{1} - 1)^{2}$.
In line 6, it defines the lower bound as $-\infty$. In line 7, it defines the upper
bound as $\infty$. In line 8, it defines a constraint $x_{1}^{2} \le 1$. If
it does not define this constraint, the automatic differentiation function
\textbf{prob2struct} will generate the incorrect gradient of the objective function
for the nonlinear least-square problem. For a nonlinear least-square problem
$\min f(x) = r(x)^{T}r(x), \; r: \; \Re^{n} \to \Re$, \textbf{prob2struct} returns
the Jacobian $J(x)$ of $r(x)$ and it is favorable to the solver of a nonlinear
least-square problem such as the Gauss-Newton method (p. 279 in \cite{NW1999})
or the Levenberg-Marquardt method (p. 282 in \cite{NW1999}).

\vskip 2mm

In line 9, the subroutine \textbf{optimproblem} assigns the objective function
and the constraints of an optimization problem. In line 10, the subroutine
\textbf{prob2struct} generates a standard MATLAB function
\textbf{generatedObjective} including the objective function and its gradient of
an optimization problem. Its gradient is computed by automatic differentiation
with the reverse mode. In line 11, it calls the CNMGE solver to find an optimal
approximation solution of the global optimization problem.

\section{Numerical experiments}

In this section, some numerical experiments are conducted to test the performance
of the continuation Newton method with the deflation technique and the quasi-genetic
evolution (CNMGE) for global optimization problems. In Subsection \ref{SUBSECVEAC},
we verify the performance improvement of the algorithm combination. Namely, we
compare the continuation method with the multi-start method (CNMTrM, Algorithm
\ref{ALGCNMTR} with the multi-start method), the continuation method with the
revised deflation technique (CNMDTM, Algorithm \ref{ALGCNMDTM}), and the
continuation method with the revised deflation technique and the evolutionary
algorithm (CNMGE, the combination of Algorithm \ref{ALGCNMDTM} and Algorithm
\ref{ALGQGE}) for 68 open test problems.

\vskip 2mm

In Subsection \ref{SUBSECTPC}, in order to test the total performance of CNMGE, we
compare it with other representative global optimization methods such as
the multi-start methods (the built-in subroutine GlobalSearch.m of MATLAB R2021b
\cite{MATLAB,ULPG2007}, GLODS: Global and Local Optimization using
Direct Search \cite{CM2015}, VRBBO: Vienna Randomized Black Box Optimization
\cite{KN2022}), the branch-and-bound method (Couenne: Convex Over- and Under-ENvelopes
for Nonlinear Estimation,  one of the state-of-the-art open-source solver
\cite{BLLMW2009,Couenne2020}), and the derivative-free algorithms
(CMA-ES \cite{Hansen2006,Hansen2010} and MCS \cite{HN1999,Neumaier2000}) for 148
open test problems from CUTEst \cite{GOT2015}.

\vskip 2mm

\subsection{Verifying the effect of the algorithm combination} \label{SUBSECVEAC}

In this subsection, in order to verify the performance improvement of the
algorithm combination, we compare the continuation method with the multi-start method
(CNMTrM, Algorithm \ref{ALGCNMTR} with the multi-start method), the continuation method
with the revised deflation technique (CNMDTM, Algorithm \ref{ALGCNMDTM}), and the
continuation method with the revised deflation technique and the evolutionary algorithm
(CNMGE, the combination of Algorithm \ref{ALGCNMDTM} and Algorithm \ref{ALGQGE}) for 68
open test problems. The test problems can be found in
\cite{Andrei2008,AD2005,EVNTBHA2019,LM2004,MGH1981,SB2020}
and their scales vary from dimension $1$ to $1000$. For problems $37, \, 60, \, 62, \, 67$,
the original test problems have the default box constraints. Since we only consider
the global optimization problems, we compute these problems without the default box
constraints.

\vskip 2mm

In order to evaluate the effect of the algorithm gradient computed by
automatic differentiation, we also compare CNMGE using the algorithm gradient computed
by automatic differentiation (we still denote it as CNMGE) with CNMGE using the analytical
gradient (we denote it as CNMGE\_AG). The termination condition of Algorithm \ref{ALGCNMTR}
for finding a stationary
point of the objective function $f(x)$ is
\begin{align}
   \|\nabla f(x_{k})\|_{\infty} \le 10^{-6}. \label{TERMCON}
\end{align}

\vskip 2mm

The codes are executed by a HP notebook with the Intel quad-core CPU and 8Gb
memory in the MATLAB2021b environment. The numerical results are
arranged in Tables \ref{TABACOMLS1}-\ref{TABACOMSS2} of Appendix \ref{APPNRCNMGET}
and Figures \ref{FIGACNMGECPUL}-\ref{FIGACNMGECPUS}. Since the global minima of
some test problems are unknown, we mark their global minima of those problems
with the notation ``$\backslash$'' in Tables \ref{TABACOMLS1}-\ref{TABACOMSS2}.
For those problems, we regard that the method fails to find their global minima
if the found minima are greater than those of other methods. The statistical
results are put in Table \ref{TABSTARCNMGE}.

\vskip 2mm

From Table \ref{TABSTARCNMGE}, we find that CNMGE and CNMGE\_AG
work well and almost obtain the global minima of 68 test problems. CNMTrM fails
to find the global minima of 27 test problems about 39.71\%. And CNMDTM fails
to find the global minima of 9 test problems about 13.23\%. Therefore, the
performance of the algorithm combination gradually becomes better and better
from the continuation Newton method with the multi-start method (CNMTrM,
Algorithm \ref{ALGCNMTR}), CNMTr with the deflation technique (CNMDTM,
Algorithm \ref{ALGCNMDTM}), and CNMDTM with the evolutionary algorithm
(CNMGE, the combination of Algorithm \ref{ALGCNMDTM} and Algorithm \ref{ALGQGE}).

\vskip 2mm

We also find that CNMGE will not sacrifice its performance if it replaces
the analytical gradient with the algorithm gradient computed by the automatic
differentiation technique with the reverse mode. We find that CNMGE and CNMGE\_AG
almost take the same time to compute a problem from Figures
\ref{FIGACNMGECPUL}-\ref{FIGACNMGECPUS}. Consequently, CNMGE with the automatic
differentiation technique retains the usability of the derivative-free method
and has the fast convergence of the gradient-based method.

\vskip 2mm

\begin{table}[htbp]
  \renewcommand{\arraystretch}{1.5}
  \newcommand{\tabincell}[2]{\begin{tabular}{@{}#1@{}}#2\end{tabular}}
  \centering
  \caption{The number of failure problems (no. 1-68) computed by CNMTrM, CNMDTM, CNMGE and CNMGE\_AG.}
  \label{TABSTARCNMGE}
  \resizebox{\textwidth}{!}{
    \begin{tabular}{|c|c|c|c|c|c|c|c|}
    \hline

    & CNMTrM & CNMDTM & CNMGE & CNMGE\_AG \cr\hline

    The number of failed problems
    & 27 & 9  & 3 & 3   \\ \hline
    The probability of failure & $\frac{27}{68} \; (39.71\%)$ &  $\frac{9}{68} \; (13.23\%)$ &
    $\frac{3}{68} \; (4.41\%)$ & $\frac{3}{68} \; (4.41\%)$ \\ \hline
    \end{tabular}}
\end{table}

\begin{figure}[htbp]
   \begin{minipage}[t]{0.49 \linewidth}
       \center
       \includegraphics[width=0.95 \textwidth]{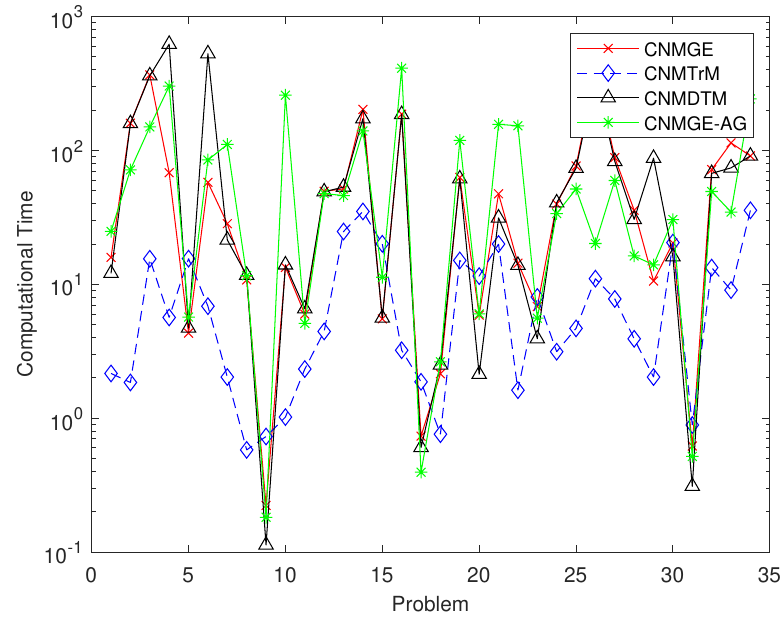}
       \caption{CPU time of large-scale problems (no. 1-34) computed by CNMGE, CNMTrM, CNMDTM and CNMGE\_AG.}
        \label{FIGACNMGECPUL}
       \end{minipage}%
       \hfill \hskip 2mm
    \begin{minipage}[t]{0.51 \linewidth}
        \centering
        \includegraphics[width=0.95 \textwidth]{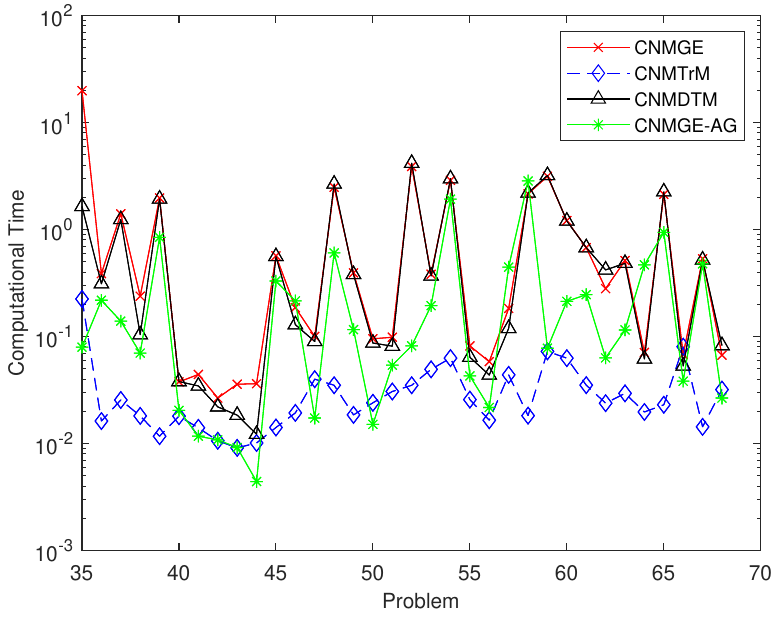}
         \caption{CPU time of small-scale problems (no. 35-68) computed by CNMGE, CNMTrM, CNMDTM and CNMGE\_AG.}
         \label{FIGACNMGECPUS}
    \end{minipage}
\end{figure}

\vskip 2mm

\subsection{Testing the total performance of CNMGE} \label{SUBSECTPC}

\vskip 2mm

In this subsection, in order to evaluate the total performance of CNMGE, we
compare it with other representative global optimization methods such as
the multi-start algorithms (the built-in subroutine GlobalSearch.m of MATLAB R2021b,
GLODS \cite{CM2015}, and VRBBO \cite{KN2022}), the branch-and-bound method (Couenne,
one of the state-of-the-art open-source solver \cite{BLLMW2009,Couenne2020}), and
the derivative-free algorithms (CMA-ES \cite{Hansen2006,Hansen2010} and MCS
\cite{HN1999,Neumaier2000}). The codes are executed by a HP notebook with the Intel
quad-core CPU and 8Gb memory except for Couenne, which is executed by the NEOS server
\cite{CMM1998,Dolan2001,GM1997,NEOS2021}.

\vskip 2mm

GlobalSearch is an efficient multi-start method for global optimization problems
and it is widely used in engineering fields. GLODS (Global and Local Optimization
using Direct Search \cite{CM2015}) and VRBBO (Vienna Randomized Black Box Optimization,
its MATLAB code can be downloaded from \cite{KN2022}) are two other representative
multi-start methods. Couenne (Convex Over- and Under-ENvelopes for Nonlinear Estimation)
is a state-of-the-art open source solver for global optimization problems.
The covariance matrix adaptation evolution strategy (CMA-ES, its MATLAB code
can be downloaded from \cite{CMAES2012}) and the multilevel coordinate search
(MCS, its MATLAB code can be downloaded from \cite{MCS2000}) are two
representative derivative-free algorithms. Therefore, we select these six
methods (GlobalSearch, GLODS, VRBBO, Couenne, CMA-ES and MCS) as the basis
for comparison.

\vskip 2mm

\textcolor{blue}{In order to test those seven methods (CNMGE, GlobalSearch,
GLODS, VRBBO, Couenne, CMA-ES and MCS) more sufficiently, we not only test
them for 68 unconstrained optimization problems from the references
\cite{Andrei2008,AD2005,EVNTBHA2019,LM2004,MGH1981,SB2020}, but also test them
for additional 80 unconstrained optimization problems from a larger publicly
available test set (CUTEst \cite{GOT2015}).} The numerical results are arranged
in Tables \ref{TABCOMLS1T}-\ref{TABCOMCUTE10} of Appendix \ref{APPNRCNMGET}.
Since the global minima of some test problems are unknown, we mark their global
minima of those problems with the notation ``$\backslash$'' in Tables
\ref{TABCOMLS1T}-\ref{TABCOMCUTE10}. For those problems, we regard that the method
fails to find their global minima if the found minima are greater than
those of other methods. The number of CMA-ES's evolution generations is set
to $10^{5}$. Due to the randomness of CMA-ES and VRBBO, we repeatedly calculate
every problem about ten times with CMA-ES and VRBBO, and select the computed
minima as their optimal values in Tables \ref{TABCOMLS1T}-\ref{TABCOMCUTE10},
respectively.

\vskip 2mm

When the unconstrained optimization problem \eqref{UNOPTF} has many local minima,
according to our numerical experiments, Algorithm \ref{ALGCNMDTM} can find most
of stationary points of $f(x)$. Thus, it makes CNMGE possibly find
the global minimum of $f(x)$ for the difficult problem. We illustrate this result
via Problem 1. Problem 1 \cite{LM2004} is a molecular potential energy problem
and its objective function has the following form:
\begin{align}
     f(\omega) = \sum_{i=1}^{n}\left(1\ +\ cos(3\omega_{i})\ +\ \frac{(-1)^{i}}
     {\sqrt{10.60099896-4.141720682(cos\omega_{i})}}\right),  \label{PEMP}
\end{align}
where $\omega_{i}$ is the torsion angle and $n+3$ is the number of atoms or beads
in the given system. The number of its local minima increases exponentially
with the size of the problem, which characterizes the principal difficult in
minimizing molecular potential energy functions for the traditional global
optimization methods. CNMGE can find 17 stationary points of problem \eqref{PEMP}
with $n = 1000$ (if CNMGE uses the Hessian matrix $H(x)$, it can find 90 stationary
points) and its global minimum $f(\omega^{\ast}) =-41.118303$ located at
$\omega^{\ast} = (1.039195, \, 3.141593, \, \ldots, \, 1.039195,\, 3.141593)$.
However, for problem 1 with 50 atoms, the traditional global
optimization methods such as the interval analysis method \cite{kearfott1996,LM2004}
or the multi-start methods (GlobalSearch, GLODS, VRBBO) are difficult to find its
global minimum. Couenne, CMA-ES and MCS also fail to solve this problem.

\vskip 2mm

\begin{figure}[htbp]
    \begin{minipage}{0.9\textwidth}
        \centering
        \includegraphics[width=0.9\linewidth]{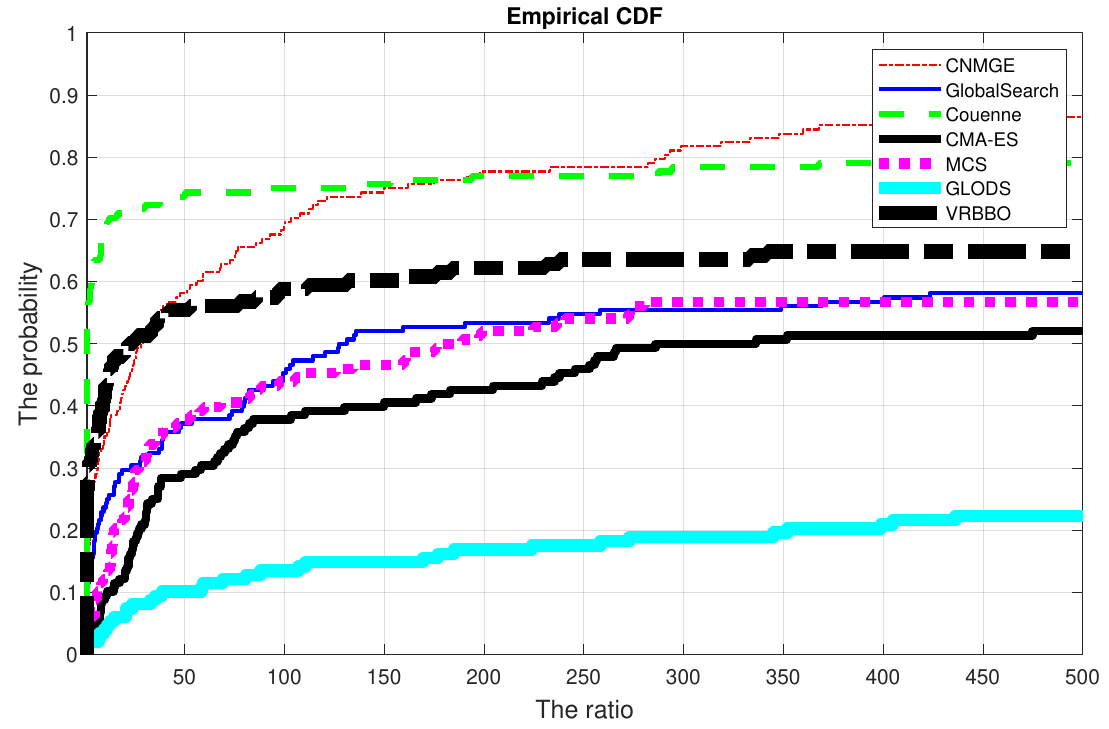}
        \captionof{figure}{Performance profile for global optimization solvers.}
        \label{FIGPCNMGE}
    \end{minipage}
\end{figure}

\textcolor{blue}{In order to evaluate and compare those seven methods (CNMGE,
GlobalSearch, Couenne, CMA-ES, MCS, GLODS and VRBBO) fairly, we also adopt the
performance profile as a evaluation tool \cite{DM2002,SKM2017}.
The performance profile for a solver is the (cumulative) distribution function
for a performance metric, which is the ratio of the computing time of the solver
versus the best time of all of the solvers as the performance metric. If the
solver fails to solve a problem, we let its ratio be a bigger number such as
$999$. Figure \ref{FIGPCNMGE} is the performance profiles for seven global
optimization solvers (CNMGE, GlobalSearch, Couenne, CMA-ES, MCS, GLODS and
VRBBO). We also count the statistic number of failed problems and fasted problems
computed by CNMGE, GlobalSearch, Couenne, CMA-ES, MCS, GLODS and
VRBBO. The statistical results are put in Table \ref{TABSRCNMGET}.}

\vskip 2mm

\textcolor{blue}{From Tables \ref{TABCOMLS1T}-\ref{TABCOMCUTE10}, Table
\ref{TABSRCNMGET} and Figure \ref{FIGPCNMGE}, we find that CNMGE works well for
unconstrained optimization problems and finds their global minima efficiently.
GlobalSearch, Couenne, CMA-ES and MCS all work well for small-scale problems.
From Table \ref{FIGPCNMGE}, we find that CNMGE fails to find the global minima
of test problems about $14/148 \; (9.46 \%)$. GlobalSearch, CMA-ES, MCS and
VRBBO fail to find the global minima of test problems about $50/148 \; (33.78 \%)$.
Couenne can solve the most problems efficiently and it can not solve efficiently
about $35/148 \; (23.65 \%)$ (Couenne fails to solve 11 test problems and 24 problems
are solved by it about 8 hours). Therefore, CNMGE is more robust than other
representative global optimization methods such as GlobalSearch, Couenne, CMA-ES,
MCS, GLODS and VRBBO.}

\vskip 2mm

\textcolor{blue}{From Table \ref{TABSRCNMGET}, we also find that Couenne is
significantly faster than the other six global optimization solvers (CNMGE,
GlobalSearch, CMA-ES, MCS, GLODS and VRBBO). One reason is due to the more
gain of the complied language than that of the interpreted language, since
Couenne is written by C++ language (a complied language), and the other six
solvers ((CNMGE, GlobalSearch, CMA-ES, MCS, GLODS and VRBBO) are written by
MATLAB language (an interpreted language). The other reason is that Couenne is
executed by a high performance (NEOS Server \cite{NEOS2021},
\url{https://neos-server.org/neos/}) and the other six solvers ((CNMGE,
GlobalSearch, CMA-ES, MCS, GLODS and VRBBO) are executed by a HP notebook with
the Intel quad-core CPU and 8Gb memory.}

\vskip 2mm

\begin{table}[htbp]
  \renewcommand{\arraystretch}{1.5}
  \newcommand{\tabincell}[2]{\begin{tabular}{@{}#1@{}}#2\end{tabular}}
  \centering
  \caption{The number of failed problems for 148 test problems computed by CNMGE, GlobalSearch,
  Couenne, CMA-ES, MCS, GLODS and VRBBO.}
  \label{TABSRCNMGET}
  \resizebox{\textwidth}{!}{
    \begin{tabular}{|c|c|c|c|c|c|c|c|}
    \hline
    & CNMGE & GlobalSearch & Couenne & CMA-ES \cr\hline
    The number of failed problems
    & 14 & 50  & {11} & 50   \cr \hline
    \tabincell{c}{The number of problems \\ computed by 8 hours}
    & $\backslash$ &  $\backslash$ & 24  & 5 \cr \hline
     {The probability of failure}
    & $\frac{14}{148} \; (9.46\%)$ & $\frac{50}{148} \; (33.78\%)$ & $\frac{35}{148}
    \; (23.65\%)$ & $\frac{55}{148} \; (37.16\%)$ \cr \hline
    The number of fasted problems
    & 13 & 13 & 72  & 3 \cr \hline
    The probability of fasted problem
    & $\frac{13}{148} \; (8.78 \%)$ & $\frac{13}{148} \; (8.78 \%)$
    & $\frac{72}{148} \; (48.65 \%)$ & $\frac{3}{148} \; (2.03 \%)$\cr \hline \hline
    & MCS & GLODS & VRBBO & $\backslash$  \cr \hline
    The number of failed problems
    & 45  & 64 & 50  &$\backslash$ \cr\hline
    \tabincell{c}{The number of problems \\ computed by 8 hours}
    & 10 &  $\backslash$ & $\backslash$  & $\backslash$ \cr \hline
    {The probability of failure}
    & $\frac{55}{148} \; (37.16\%)$ & $\frac{64}{148} \; (43.24\%)$ & $\frac{50}{148}
    \; (33.78\%)$ & $\backslash$  \cr \hline
    The number of fasted problems
    & 5 & 2 & 40  & \cr \hline
    The probability of fasted problem
    & $\frac{5}{148} \; (3.38 \%)$ & $\frac{2}{148} \; (1.35 \%)$
    & $\frac{40}{148} \; (27.03 \%)$ & \cr \hline \hline
  \end{tabular}}
\end{table}

\vskip 2mm

\vskip 2mm

\section{Conclusions}

\vskip 2mm

Based on the deflation technique and the genetic evolution, we give an efficient
continuation Newton for solving the global minimum of an unconstrained
optimization problem. We implement it in the MATLAB2021b environment and denote
this solver as CNMGE. In order to retain the usability of derivative-free methods
and the fast convergence of the gradient-based methods, we use the automatic
differentiation technique with the reverse mode to compute the gradient
of the objective function and replace the Hessian matrix with its finite
difference approximation for CNMGE. The numerical results show that CNMGE
works well for global optimization problems, especially some large-scale
problems of which are difficult to be solved by the known global optimization
methods. Therefore, CNMGE can be regarded as an alternative workhorse for
unconstrained optimization problems.

\vskip 2mm

\textcolor{blue} {We only compare CNMGE with some state-of-the-art open source
methods. And we do not compare CNMGE with commercial solvers
such as BARON \cite{Sahinidis2021} or non-open source solvers. In the future,
it is worthy to compare CNMGE with some no-open source state-of-the-art
methods such as the methods, which are proposed by Y.~D. Sergeyev and
D.~E. Kvasov \cite{KS2012,SK2015}, or the commercial solvers such as BARON,
which is written by N.~V. Sahinidis \cite{Sahinidis2021}. Furthermore, CNMGE
is worthy to be explored further for non-smooth optimization problems by
combining the technique of estimating the Lipschitz constant
\cite{KS2012,SK2017,SK2015}. And the computational efficiency of CNMGE can be
improved further if its parameters are selected elaborately or it combines the
state-of-the-art acceleration technique \cite{BRS2018}.}

\vskip 2mm

\section*{Acknowledgments}
The authors are grateful to Prof. Jonathan Eckstein for the suggestion of the
comparison to Couenne and Prof. Nick Sahinidis for the suggestion of the comparison
to the derivative-free optimization methods such as MCS and CMA-ES.
The authors are grateful to three anonymous referees for their comments and
suggestions which greatly improve the presentation of this paper.

\section * {Declarations}

\vskip 2mm

\noindent \textbf{Funding:} This work was supported in part by Grants 61876199 and
62376036 from National Natural Science Foundation of China, Grant YBWL2011085 from
Huawei Technologies Co., Ltd., and Grant YJCB2011003HI from the Innovation
Research Program of Huawei Technologies Co., Ltd..

\vskip 2mm

\noindent \textbf{Conflicts of interest / Competing interests:} Not applicable.

\vskip 2mm

\noindent \textbf{Ethical Approval:} Not applicable.

\vskip 2mm

\noindent \textbf{Availability of data and material (data transparency):} If it is requested, we will
provide the test data.

\vskip 2mm

\noindent \textbf{Code availability (software application or custom code):}
\url{https://github.com/luoxinlongroger/CNMGE/} or
\url{https://teacher.bupt.edu.cn/luoxinlong/zh_CN/zzcg/41406/list/index.htm}.

\begin{appendix}

\section{Tables of Numerical Results} \label{APPNRCNMGET}

\vskip 2mm

\begin{table}[htbp]
  \renewcommand{\arraystretch}{1.2}
  \centering
  \caption{Numerical results of large-scale problems (no. 1-17) computed by
  CNMGE, CNMTrM, CNMDTM and CNMGE\_AG.}
  \label{TABACOMLS1}
  \resizebox{\textwidth}{!}{
\begin{tabular}{|c|c|c|c|c|c|c|c|c|}
    \hline
     &
    \multicolumn{1}{c|}{CNMTrM} & \multicolumn{1}{c|}{CNMDTM} & \multicolumn{1}{c|}{CNMGE}
    & \multicolumn{1}{c|}{CNMGE\_AG} &
      \cr \cline{2-5}
    Problem & $\min f(x)$  & $\min f(x)$  & $\min f(x)$   &  $\min f(x)$ & {Global minimum} \cr
    & (CPU time (s)) & (CPU time (s)) &  (CPU time (s)) & (CPU time (s)) &  \cr    \hline
    1. Molecular Energy
    & \textcolor{red}{-2.8789E-01} & -41.1183E+00 & -41.1183E+00 & -41.1183E+00  &  \cr
    Problem (n = 1000) \cite{LM2004}
    &\textcolor{red}{(2.16E+00)} & (1.21E+01) & (1.59E+01) & (2.49E+01) & -4.1118E+01 \cr
    & \textcolor{red}{(failed)} & & & & \cr\hline
    2. Ackley Function
    & \textcolor{red}{3.5744E+0} &\textcolor{red}{3.5744E+00} & 8.8817E-16 & 8.8817E-16 & \cr
    (n = 1000) \cite{SB2020}
    &\textcolor{red}{(1.85E+0)}  &\textcolor{red}{(1.59E+02)}  & (1.60E+02) &(71.79) & 0 \cr
\hline
    \end{tabular}}
\end{table}

\newpage

\begin{table}[htbp]
  \renewcommand{\arraystretch}{1.5}
  \centering
  \caption{Numerical results of large-scale problems (no. 18-34) computed by
  CNMGE, CNMTrM, CNMDTM and CNMGE\_AG.}
  \label{TABACOMLS2}
  \resizebox{\textwidth}{!}{
    \begin{tabular}{|c|c|c|c|c|c|c|c|c|}
    \hline
    &
    \multicolumn{1}{c|}{CNMTrM} & \multicolumn{1}{c|}{CNMDTM} &
    \multicolumn{1}{c|}{CNMGE} & \multicolumn{1}{c|}{CNMGE\_AG} &
     \cr \cline{2-5}
    Problem & $\min f(x)$  & $\min f(x)$  & $\min f(x)$   &  $\min f(x)$ & Global minimum \cr
    & (CPU time (s)) & (CPU time (s)) &  (CPU time (s)) & (CPU time (s)) &  \cr \hline
    18. Raydan 2 Function
    & 1.0000E+03 & 1.0000E+03 & 1.0000E+03 & 1.0000E+03 &   \cr
    (n = 1000) \cite{Andrei2008}
    & (7.61E-01) & (2.50E+00) & (2.15E+00) & (2.64E+00) & $\backslash$ \cr\hline
    19. Extended Tridiagonal 1
    & 1.6699E-12 & 5.9607E-19 & 5.9698E-19 & 5.9598E-19 &   \cr
    Function(n = 1000) \cite{Andrei2008}
    & (1.51E+01) & (6.15E+01) & (6.29E+01) & (1.19E+02) & $\backslash$  \cr\hline
    20. Extended quadratic penalty
    & 3.9900E+03 & 3.9900E+03 & 3.9900E+03 & 3.9900E+03 &   \cr
\hline
    \end{tabular}}
\end{table}

\vskip 2mm

\begin{table}[htbp]
  \renewcommand{\arraystretch}{1.5}
  \centering
  \caption{Numerical results of small-scale problems (no. 35-50) computed by
  CNMGE, CNMTrM, CNMDTM and CNMGE\_AG.}
  \label{TABACOMSS1}
  \resizebox{\textwidth}{!}{
    \begin{tabular}{|c|c|c|c|c|c|c|c|c|}
    \hline
    &
    \multicolumn{1}{c|}{CNMTrM} & \multicolumn{1}{c|}{CNMDTM} & \multicolumn{1}{c|}{CNMGE}
    & \multicolumn{1}{c|}{CNMGE\_AG} &
      \cr \cline{2-5}
    {Problem} & $\min f(x)$  & $\min f(x)$  & $\min f(x)$   &  $\min f(x)$ & {Global minimum}\cr
    & (CPU time (s)) & (CPU time (s)) &  (CPU time (s)) & (CPU time (s)) &  \cr \hline
    35. Griewank Function
    & \textcolor{red}{1.0030E+00} & 2.7755E-15 & 0 & 0 &  \cr
    (n = 10) \cite{MGH1981,SB2020}
    & \textcolor{red}{(2.24E-01)} & (1.64E+00) & (1.98E+00) & (1.95E-02) & 0  \cr
    & \textcolor{red}{(failed)} & & & & \cr\hline
    36. Levy Function N. 13
    & \textcolor{red}{9.7371E-01} & 5.9152E-22 & 1.3497E-31 & 1.3497E-31 &   \cr
    (n = 2) \cite{SB2020}
    &\textcolor{red}{(1.62E-02)} & (3.11E-01) & (3.87E-01) & (2.18E-01) & 0 \cr
\hline
    \end{tabular}}
\end{table}

\vskip 2mm

\begin{table}[htbp]
  \renewcommand{\arraystretch}{1.5}
  \centering
  \caption{Numerical results of small-scale problems (no. 51-68) computed by
  CNMGE, CNMTrM, CNMDTM and CNMGE\_AG.}
  \label{TABACOMSS2}
  \resizebox{\textwidth}{!}{
    \begin{tabular}{|c|c|c|c|c|c|c|c|c|}
    \hline
    &
    \multicolumn{1}{c|}{CNMTrM} & \multicolumn{1}{c|}{CNMDTM} &
    \multicolumn{1}{c|}{CNMGE} & \multicolumn{1}{c|}{CNMGE\_AG} &
     \cr \cline{2-5}
    Problem & $\min f(x)$  & $\min f(x)$  & $\min f(x)$   &  $\min f(x)$ & Global minimum \cr
    & (CPU time (s)) & (CPU time (s)) &  (CPU time (s)) & (CPU time (s)) &  \cr    \hline
    51. Six-Hump Camel
    & \textcolor{red}{5.4371E-01} & -1.0316E+00 & -1.0316E+00 & -1.0316E+00 &   \cr
    Function (n = 2) \cite{SB2020}
    & \textcolor{red}{(3.06E-02)} & (8.12E-02) & (9.84E-02) & (5.39E-02) &  -1.0316E+00 \cr
    &\textcolor{red}{(failed)} & & & & \cr\hline
    52. Drop-Wave Function
    & \textcolor{red}{-8.6245E-15} & -1.0000E+00 & -1.0000E+00 & -1.0000E+00 &  \cr
    (n = 2) \cite{SB2020}
    & \textcolor{red}{(3.50E-02)} & (4.16E+00) & (3.88E+00) & (8.18E-02) &  -1.0000E+00 \cr
\hline
    \end{tabular}}
\end{table}

\vskip 2mm

\begin{table}[htbp]
  \renewcommand{\arraystretch}{1.}
  \centering
  \caption{Numerical results of large-scale problems (no.1-17) computed
  by CNMGE, GlobalSearch, Couenne, and CMA-ES.}
  \label{TABCOMLS1T}
  \resizebox{\textwidth}{!}{
    \begin{tabular}{|c|c|c|c|c|c|c|c|c|}
    \hline
    &
    \multicolumn{1}{c|}{CNMGE} & \multicolumn{1}{c|}{GlobalSearch} & \multicolumn{1}{c|}{Couenne}
    & \multicolumn{1}{c|}{CMA-ES} &  \cr \cline{2-5}
    Problem & $\min f(x)$  & $\min f(x)$  & $\min f(x)$   &  $\min f(x)$ & Global minimum \cr
    & (CPU time (s)) & (CPU time (s)) &  (CPU time (s)) & (CPU time (s)) &  \cr    \hline
    1. Molecular Energy
    & -4.1118E+01 &\textcolor{red}{-2.8761E-01} & -4.1118E+01 & \textcolor{red}{1.0008E+01}  &     \cr
    Problem (n = 1000) \cite{LM2004}
    & (1.59E+01)  & \textcolor{red}{(1.82E+01)}  & (2.68E+04) &  \textcolor{red}{(2.98E+02)}
    & -4.1118E+01 \cr
    & & \textcolor{red}{(failed)}& \textcolor{red}{(too long)} & \textcolor{red}{(failed)}
    &  \cr\hline
    2. Ackley Function
    & 8.8817E-016 &8.8817E-16 & 6.7078E-06  & \textcolor{red}{3.6253}  &  \cr
    (n = 1000) \cite{SB2020}
    & (1.60E+02)  & (1.81E+01)  & (1.77E+03) & \textcolor{red}{(7.71E+01)}  & 0 \cr
\hline
    \end{tabular}}
\end{table}

\newpage

\begin{table}[htbp]
  \renewcommand{\arraystretch}{1.}
  \centering
  \caption{Numerical results of large-scale problems (no.18-34) computed
  by CNMGE, GlobalSearch, Couenne and CMA-ES.}
  \label{TABCOMLS2T}
  \resizebox{\textwidth}{!}{
    \begin{tabular}{|c|c|c|c|c|c|c|c|c|}
    \hline
    &
    \multicolumn{1}{c|}{CNMGE} & \multicolumn{1}{c|}{GlobalSearch} & \multicolumn{1}{c|}{Couenne}
    & \multicolumn{1}{c|}{CMA-ES} & \cr \cline{2-5}
    Problem & $\min f(x)$  & $\min f(x)$  & $\min f(x)$   &  $\min f(x)$ & Global minimum \cr
    & (CPU time (s)) & (CPU time (s)) &  (CPU time (s)) & (CPU time (s)) &  \cr    \hline
    18. Raydan 2 Function
    & 1.0000E+03 & 1.0000E+03  & 1.0000E+03 & \textcolor{red}{1.3083E+03} &  \cr
    (n = 1000) \cite{Andrei2008}
    & (2.15E+00)  & (7.60E+00)  & (9.68E-02) & \textcolor{red}{(2.93E+02)}  & $\backslash$\cr
      & & & & \textcolor{red}{(failed)} & \cr\hline
    19. Extended Tridiagonal 1
    & 5.9698E-19 &\textcolor{red}{2.9700E+02}  & \textcolor{red}{9.9721E+02} &
    \textcolor{red}{1.0000E+03} &  \cr
    Function(n = 1000) \cite{Andrei2008}
    & (6.29E+01)  & \textcolor{red}{(4.51E+00)} & \textcolor{red}{(5.61E+02)}  &
    \textcolor{red}{(2.66E+02)} & $\backslash$\cr
\hline
    \end{tabular}}
\end{table}

\vskip 2mm

\begin{table}[htbp]
  \renewcommand{\arraystretch}{1.}
  \centering
  \caption{Numerical results of small-scale problems (no.35-51) computed
  by CNMGE, GlobalSearch, Couenne and CMA-ES.}
  \label{TABCOMSS1T}
  \resizebox{\textwidth}{!}{
    \begin{tabular}{|c|c|c|c|c|c|c|c|c|}
    \hline
    &
    \multicolumn{1}{c|}{CNMGE} & \multicolumn{1}{c|}{GlobalSearch} & \multicolumn{1}{c|}{Couenne}
    & \multicolumn{1}{c|}{CMA-ES} &  \cr \cline{2-5}
    Problem & $\min f(x)$  & $\min f(x)$  & $\min f(x)$   &  $\min f(x)$ & Global minimum \cr
    & (CPU time (s)) & (CPU time (s)) &  (CPU time (s)) & (CPU time (s)) &  \cr \hline
    35. Griewank Function
    & 1.9318E-14 & 1.7187E-12  & 1.5929E-26 & 6.6613E-16  &  \cr
    (n = 10) \cite{MGH1981,SB2020}
    & (1.98E+00)  & (2.66E+00)  & (1.14E-03) & (1.33E+00)  & 0  \cr \hline
    36. Levy Function N. 13
    & 1.3497E-31 &\textcolor{red}{1.0981E-01} & 2.8774E-16  & 1.3498E-31 & \cr
    (n = 2) \cite{SB2020}
    & (3.87E-01)  & \textcolor{red}{(2.85E+00)}  & (8.48E-03) &  (6.78E-01) & 0 \cr
      & & \textcolor{red}{(failed)} & & & \cr
\hline
    \end{tabular}}
\end{table}

\vskip 2mm

\begin{table}[htbp]
  \renewcommand{\arraystretch}{1.}
  \centering
  \caption{Numerical results of small-scale problems (no.52-68) computed
  by CNMGE, GlobalSearch, Couenne and CMA-ES.}
  \label{TABCOMSS2T}
  \resizebox{\textwidth}{!}{
    \begin{tabular}{|c|c|c|c|c|c|c|c|c|}
    \hline
    &
    \multicolumn{1}{c|}{CNMGE} & \multicolumn{1}{c|}{GlobalSearch} & \multicolumn{1}{c|}{Couenne}
    & \multicolumn{1}{c|}{CMA-ES} &   \cr \cline{2-5}
    Problem & $\min f(x)$  & $\min f(x)$  & $\min f(x)$   &  $\min f(x)$ & Global minimum \cr
    & (CPU time (s)) & (CPU time (s)) &  (CPU time (s)) & (CPU time (s)) &  \cr \hline
    52. Drop-Wave Function
    & -1.0000E+00 & -1.0000E+00 & -1.0000E+00 & {-9.8389E-01} &   \cr
    (n = 2) \cite{SB2020}
    & (3.88E+00)  & (3.12E+00) & (2.59E-02) & {(7.34E-01)} & -1.0000E+00 \cr\hline
    53. Perm Function 0, d, $\beta$
    & 0 &\textcolor{red}{6.3025E+02} & 5.3868E-10 & 3.0872E-16 &  \cr
    (n = 4) \cite{SB2020}
    & (3.82E-01)  & \textcolor{red}{(5.79E+00)} & (2.10E-01)  &  (1.11E+00)  & 0 \cr
      & & \textcolor{red}{(failed)} & & & \cr
\hline
    \end{tabular}}
\end{table}

\vskip 2mm

\begin{table}[htbp]
  \renewcommand{\arraystretch}{1.}
  \centering
  \caption{Numerical results of large-scale problems (no.1-17) computed by
  CNMGE, MCS, GLODS and VRBBO.}
  \label{TABCOMLS3T}
  \resizebox{\textwidth}{!}{
    \begin{tabular}{|c|c|c|c|c|c|c|c|c|}
    \hline
    &
    \multicolumn{1}{c|}{CNMGE} & \multicolumn{1}{c|}{MCS} & \multicolumn{1}{c|}{GLODS}
    & \multicolumn{1}{c|}{VRBBO} &   \cr \cline{2-5}
    Problem & $\min f(x)$  & $\min f(x)$  & $\min f(x)$   &  $\min f(x)$ & Global minimum \cr
    & (CPU time (s)) & (CPU time (s)) &  (CPU time (s)) & (CPU time (s)) &  \cr \hline
    1. Molecular Energy
    & -4.1118E+01 &\textcolor{red}{-3.6903E-01} & \textcolor{red}{6.2332E-04} &
    \textcolor{red}{-2.8789E-01}  &  \cr
    Problem (n = 1000) \cite{LM2004}
    & (1.59E+01)  & \textcolor{red}{(2.01E+02)}  & \textcolor{red}{(1.21E+02)} &
    \textcolor{red}{(1.51E+01)} & -4.1118E+01  \cr
    & & \textcolor{red}{(failed)}& \textcolor{red}{(failed)} &
    \textcolor{red}{(failed)}  &  \cr \hline
    2. Ackley Function
    & 8.8817E-16 &0 & 8.8817E-16  &4.8252E-07  &  \cr
    (n = 1000) \cite{SB2020}
    & (1.60E+02)  & (1.28E+04)  & (1.14E+02) & (4.81E+00)  & 0 \cr
\hline
    \end{tabular}}
\end{table}

\vskip 2mm

\begin{table}[htbp]
  \renewcommand{\arraystretch}{1.}
  \centering
  \caption{Numerical results of large-scale problems (no.18-34) computed by
  CNMGE, MCS, GLODS and VRBBO.}
  \label{TABCOMLS4T}
  \resizebox{\textwidth}{!}{
    \begin{tabular}{|c|c|c|c|c|c|c|c|c|}
    \hline
    &
    \multicolumn{1}{c|}{CNMGE} & \multicolumn{1}{c|}{MCS} & \multicolumn{1}{c|}{GLODS}
    & \multicolumn{1}{c|}{VRBBO} &   \cr \cline{2-5}
    Problem & $\min f(x)$  & $\min f(x)$  & $\min f(x)$   &  $\min f(x)$ & Global minimum \cr
    & (CPU time (s)) & (CPU time (s)) &  (CPU time (s)) & (CPU time (s)) &  \cr    \hline
    18. Raydan 2 Function
    & 1.0000E+03 & 1.0000E+03  & 1.0000E+03 & 1.3083E+03 &  \cr
    (n = 1000) \cite{Andrei2008}
    & (2.15E+00)  & (1.31E+04)  & (5.16E+01) & (5.08E+00)  & $\backslash$ \cr
    & & \textcolor{red}{(too long)} & & & \cr \hline
    19. Extended Tridiagonal 1
    & 5.9698E-19 &\textcolor{red}{4.0306E-01}  & \textcolor{red}{9.8000E+02} &
    9.0949E-13 &  \cr
    Function(n = 1000) \cite{Andrei2008}
    & (6.29E+01)  & \textcolor{red}{(1.59E+02)} & \textcolor{red}{(1.47E+02)}  &
    (4.76E+00)   & $\backslash$\cr
\hline
  \end{tabular}}
\end{table}

\vskip 2mm

\begin{table}[htbp]
  \renewcommand{\arraystretch}{1.}
  \centering
  \caption{Numerical results of small-scale problems (no.35-51) computed by
  CNMGE, MCS, GLODS and VRBBO.}
  \label{TABCOMSS3T}
  \resizebox{\textwidth}{!}{
    \begin{tabular}{|c|c|c|c|c|c|c|c|c|}
    \hline
    &
    \multicolumn{1}{c|}{CNMGE} & \multicolumn{1}{c|}{MCS} & \multicolumn{1}{c|}{GLODS}
    & \multicolumn{1}{c|}{VRBBO} &  \cr \cline{2-5}
    Problem & $\min f(x)$  & $\min f(x)$  & $\min f(x)$   &  $\min f(x)$ & Global minimum \cr
    & (CPU time (s)) & (CPU time (s)) &  (CPU time (s)) & (CPU time (s)) &  \cr \hline
    35. Griewank Function
    & 1.9318E-14 & 0  & 0 & 3.2362E-09  &  \cr
    (n = 10) \cite{MGH1981,SB2020}
    & (1.98E+00)  & (6.38E-01)  & (9.01E+00) & (2.63E-01)  & 0 \cr \hline
    36. Levy Function N. 13
    & 1.3497E-31 & 1.3498E-31 & 1.3497E-31  & 5.4136E-06 & \cr
    (n = 2) \cite{SB2020}
    & (3.87E-01)  & (7.01E-01)  & (5.80E+01) &  (3.74E-02) & 0 \cr \hline
    37. Hosaki Function
    & -$\infty$ & -2.3457E+00 & -2.3458E+00 & -1.0000E+50 &  \cr
\hline
  \end{tabular}}
\end{table}

\vskip 2mm

\begin{table}[htbp]
  \renewcommand{\arraystretch}{1.}
  \centering
  \caption{Numerical results of small-scale problems (no.52-68) computed by
  CNMGE, MCS, GLODS and VRBBO.}
  \label{TABCOMSS4T}
  \resizebox{\textwidth}{!}{
    \begin{tabular}{|c|c|c|c|c|c|c|c|c|}
    \hline
    &
    \multicolumn{1}{c|}{CNMGE} & \multicolumn{1}{c|}{MCS} & \multicolumn{1}{c|}{GLODS}
    & \multicolumn{1}{c|}{VRBBO} &  \cr \cline{2-5}
    Problem & $\min f(x)$  & $\min f(x)$  & $\min f(x)$   &  $\min f(x)$ & Global Minimum \cr
    & (CPU time (s)) & (CPU time (s)) &  (CPU time (s)) & (CPU time (s)) &  \cr \hline
    52. Drop-Wave Function
    & -1.0000E+00 & -1.0000E+00 & -1.0000E+00 & {-7.8506E-01} &   \cr
    (n = 2) \cite{SB2020}
    & (3.88E+00)  & (5.65E-01) & (5.32E+02) & {(1.74E-01)} &
    -1.0000E+00 \cr \hline
    53. Perm Function 0, d, $\beta$
    & 0 &\textcolor{red}{1.0130E+02} & \textcolor{red}{1.3286E+00} &
    \textcolor{red}{7.3526E+02} &  \cr
    (n = 4) \cite{SB2020}
    & (3.82E-01)  & \textcolor{red}{(9.28E-01)} & \textcolor{red}{(2.16E+01)}  &
    \textcolor{red}{(4.43E-02)} & 0 \cr
    & & \textcolor{red}{(failed)} &\textcolor{red}{(failed)} &
    \textcolor{red}{(failed)} & \cr
\hline
  \end{tabular}}
\end{table}

\vskip 2mm

\begin{table}[htbp]
  \renewcommand{\arraystretch}{1.}
  \centering
  \caption{Numerical results of CUTEst problems \cite{GOT2015} (no. 69-84) computed
  by CNMGE, GlobalSearch, Couenne, and CMA-ES.}
  \label{TABCOMCUTE1}
  \resizebox{\textwidth}{!}{
    \begin{tabular}{|c|c|c|c|c|c|c|c|c|}
    \hline
    &
    \multicolumn{1}{c|}{CNMGE} & \multicolumn{1}{c|}{GlobalSearch} & \multicolumn{1}{c|}{Couenne}
    & \multicolumn{1}{c|}{CMA-ES} &  \cr \cline{2-5}
    Problem & $\min f(x)$  & $\min f(x)$  & $\min f(x)$   &  $\min f(x)$ & Global minimum \cr
    & (CPU time (s)) & (CPU time (s)) &  (CPU time (s)) & (CPU time (s)) &  \cr    \hline
    69. Allinitu
    & 5.7444 & 5.7444 & 5.7444 & 5.7444 &  \cr
    (n = 4) & (4.50E-01) & (2.19E+00) & (4.36E-03) & (2.07E+00) & $\backslash$
    \cr \hline
    70. cliff
    & 1.9979E-01 & 1.9979E-01 & 2.0724E-01 & 1.9992E-01  &  \cr
    (n = 2)  & (2.63E+00)   & (3.78E-01)  & (4.72E-03)  & (1.66E+00)  &
    $\backslash$ \cr \hline
    71. dixon3dq
    & 0 & 4.8186E-14 & 2.2326E-32 & 3.6187E-16 &  \cr
\hline
  \end{tabular}}
\end{table}

\vskip 2mm

\begin{table}[htbp]
  \renewcommand{\arraystretch}{1.}
  \centering
  \caption{Numerical results of CUTEst problems \cite{GOT2015} (no.85-100) computed
  by CNMGE, GlobalSearch, Couenne, and CMA-ES.}
  \label{TABCOMCUTE2}
  \resizebox{\textwidth}{!}{
    \begin{tabular}{|c|c|c|c|c|c|c|c|c|}
    \hline
    &
    \multicolumn{1}{c|}{CNMGE} & \multicolumn{1}{c|}{GlobalSearch} & \multicolumn{1}{c|}{Couenne}
    & \multicolumn{1}{c|}{CMA-ES} &  \cr \cline{2-5}
    Problem & $\min f(x)$  & $\min f(x)$  & $\min f(x)$   &  $\min f(x)$ & Global minimum \cr
    & (CPU time (s)) & (CPU time (s)) &  (CPU time (s)) & (CPU time (s)) &  \cr \hline
    85. nondquar
    & 0.0000E+00  & \textcolor{red}{4.8650E+03} & 4.1323E-11 &
    \textcolor{red}{8.4830E+04} &  \cr
    (n = 1000)  & (1.04E+02) & \textcolor{red}{(2.32E+00)} & (2.89E-01) &
    \textcolor{red}{(1.83E+02)}
    & $\backslash$  \cr
    & & \textcolor{red}{(failed)}  &  &\textcolor{red}{(failed)} & \cr \hline
    86. penalty1
    & 2.5000E-03  & \textcolor{red}{6.2500E+04} & \textcolor{red}{6.4395E+00} &
    \textcolor{red}{1.2410E-02} & \cr
    (n = 1000) & (6.63E+01) & \textcolor{red}{(1.12E+00)} & \textcolor{red}{(3.75E+01)}
    & \textcolor{red}{(9.04E+01)} & $\backslash$  \cr
\hline
  \end{tabular}}
\end{table}

\vskip 2mm

\begin{table}[htbp]
  \renewcommand{\arraystretch}{1.}
  \centering
  \caption{Numerical results of CUTEst problems \cite{GOT2015} (no.101-116) computed
  by CNMGE, GlobalSearch, Couenne, and CMA-ES.}
  \label{TABCOMCUTE3}
  \resizebox{\textwidth}{!}{
    \begin{tabular}{|c|c|c|c|c|c|c|c|c|}
    \hline
    &
    \multicolumn{1}{c|}{CNMGE} & \multicolumn{1}{c|}{GlobalSearch} & \multicolumn{1}{c|}{Couenne}
    & \multicolumn{1}{c|}{CMA-ES} &  \cr \cline{2-5}
    Problem & $\min f(x)$  & $\min f(x)$  & $\min f(x)$   &  $\min f(x)$ & Global minimum \cr
    & (CPU time (s)) & (CPU time (s)) &  (CPU time (s)) & (CPU time (s)) &  \cr \hline
    101. chainwoo
    & 1.0000E+00  & 1.0000E+00  & \textcolor{red}{8.1471E+00} &
    \textcolor{red}{2.1209E+03} &  \cr
    (n = 1000)  & (9.19E+01) & (3.26E+00) & \textcolor{red}{(3.35E-01)} &
    \textcolor{red}{(1.11E+02)}
    & $\backslash$  \cr
    &   &  &\textcolor{red}{(failed)}  &\textcolor{red}{(failed)} & \cr \hline
    102. chnrosnb
    & 0  & 0 & 1.2858E-16  & 3.5330E-15  &  \cr
    (n = 50) & (2.29E+00) & (8.04E-01) & (8.12E-03) & (9.41E+00) & $\backslash$
    \cr
\hline
  \end{tabular}}
\end{table}

\vskip 2mm

\begin{table}[htbp]
  \renewcommand{\arraystretch}{1.}
  \centering
  \caption{Numerical results of CUTEst problems \cite{GOT2015} (no.117-132) computed
  by CNMGE, GlobalSearch, Couenne, and CMA-ES.}
  \label{TABCOMCUTE4}
  \resizebox{\textwidth}{!}{
    \begin{tabular}{|c|c|c|c|c|c|c|c|c|}
    \hline
    &
    \multicolumn{1}{c|}{CNMGE} & \multicolumn{1}{c|}{GlobalSearch} & \multicolumn{1}{c|}{Couenne}
    & \multicolumn{1}{c|}{CMA-ES} &  \cr \cline{2-5}
    Problem & $\min f(x)$  & $\min f(x)$  & $\min f(x)$   &  $\min f(x)$ & Global minimum \cr
    & (CPU time (s)) & (CPU time (s)) &  (CPU time (s)) & (CPU time (s)) &  \cr \hline
    117. hilbertb
    & 0  & 2.8783E-13 & 3.0878E-29  & 6.3562E-16  &  \cr
    (n = 50) & (4.14E-01) & (3.55E-01) & (2.82E+04) & (3.89E+00) & $
    \backslash$  \cr
    &  & & \textcolor{red}{(too long)} &   & \cr \hline
    118. himmelbh
    & -1.0000E+10  & -2.6987E+25  & \textcolor{red}{-1.0000E+00} &
    \textcolor{red}{-1.0000E+00} & \cr
    (n = 2) & (1.75E-01) & (2.08E+00) & \textcolor{red}{(2.81E-01)}
    & \textcolor{red}{(1.19E+00)} & $\backslash$  \cr
\hline
  \end{tabular}}
\end{table}

\vskip 2mm

\begin{table}[htbp]
  \renewcommand{\arraystretch}{1.5}
  \centering
  \caption{Numerical results of CUTEst problems \cite{GOT2015} (no.133-148) computed
  by CNMGE, GlobalSearch, Couenne, and CMA-ES.}
  \label{TABCOMCUTE5}
  \resizebox{\textwidth}{!}{
    \begin{tabular}{|c|c|c|c|c|c|c|c|c|}
    \hline
    &
    \multicolumn{1}{c|}{CNMGE} & \multicolumn{1}{c|}{GlobalSearch} & \multicolumn{1}{c|}{Couenne}
    & \multicolumn{1}{c|}{CMA-ES} &  \cr \cline{2-5}
    Problem & $\min f(x)$  & $\min f(x)$  & $\min f(x)$   &  $\min f(x)$ & Global minimum \cr
    & (CPU time (s)) & (CPU time (s)) &  (CPU time (s)) & (CPU time (s)) &  \cr \hline
    133. vardim
    & 0 & 0  & 2.4591E-09 & \textcolor{red}{1.16E-02} &  \cr
    (n = 100) & (1.46E-01) & (1.85E+00) & (1.16E-02) & \textcolor{red}{(1.03E+01)}
    & $\backslash$ \cr
    &  &  &  & \textcolor{red}{(failed)} & \cr \hline
    134. watson
    & 1.8000E-07 & 2.4192E-08 & 1.0331E-13 & 2.0126E-08 &  \cr
    (n = 31) & (7.18E+00) & (1.42E+00) & (1.36E-01) & (3.20E+01) & $\backslash$
    \cr
\hline
  \end{tabular}}
\end{table}

\vskip 2mm

\begin{table}[htbp]
  \renewcommand{\arraystretch}{1.5}
  \centering
  \caption{Numerical results of CUTEst problems \cite{GOT2015} (no.69-84) computed
  by CNMGE, MCS, GLODS, and VRBBO.}
  \label{TABCOMCUTE6}
  \resizebox{\textwidth}{!}{
    \begin{tabular}{|c|c|c|c|c|c|c|c|c|}
    \hline
    &
    \multicolumn{1}{c|}{CNMGE} & \multicolumn{1}{c|}{MCS} & \multicolumn{1}{c|}{GLODS}
    & \multicolumn{1}{c|}{VRBBO} &  \cr \cline{2-5}
    Problem & $\min f(x)$  & $\min f(x)$  & $\min f(x)$   &  $\min f(x)$ & Global minimum \cr
    & (CPU time (s)) & (CPU time (s)) &  (CPU time (s)) & (CPU time (s)) &  \cr  \hline
    69. Allinitu
    & 5.7444E+00 & 5.7444E+00 & 5.7444E+00 & 5.7444E+00 &  \cr
    (n = 4) & (4.50E-01) & (5.48E-01) & (2.83E+01) & (1.43E-01) & $\backslash$
    \cr \hline
    70. cliff
    & 1.9979E-01 & 1.8377E-01 & 2.0026E-01 &\textcolor{red}{1.5985E+02} &  \cr
    (n = 2)  & (2.63E+00) & (1.84E-01) & (5.37E+01) &\textcolor{red}{(5.82E-02)}
    & $\backslash$  \cr
    &  &  &  & \textcolor{red}{(failed)} & \cr
\hline
  \end{tabular}}
\end{table}

\vskip 2mm

\begin{table}[htbp]
  \renewcommand{\arraystretch}{1.5}
  \centering
  \caption{Numerical results of CUTEst problems \cite{GOT2015} (no.85-100) computed
  by CNMGE, MCS, GLODS, and VRBBO.}
  \label{TABCOMCUTE7}
  \resizebox{\textwidth}{!}{
    \begin{tabular}{|c|c|c|c|c|c|c|c|c|}
    \hline
    &
    \multicolumn{1}{c|}{CNMGE} & \multicolumn{1}{c|}{MCS} & \multicolumn{1}{c|}{GLODS}
    & \multicolumn{1}{c|}{VRBBO} &  \cr \cline{2-5}
    Problem & $\min f(x)$  & $\min f(x)$  & $\min f(x)$   &  $\min f(x)$ & Global minimum \cr
    & (CPU time (s)) & (CPU time (s)) &  (CPU time (s)) & (CPU time (s)) &  \cr \hline
    85. nondquar
    & 0 & 0 & 0 & 2.6618E-04 &  \cr
    (n = 1000) & (1.04E+02) & (7.74E+03) & (1.17E+02) & (2.23E+01) & $\backslash$
    \cr \hline
    86. penalty1
    & 2.5000E-03 & 4.4289E-03 & \textcolor{red}{1.2020E-02} &
    \textcolor{red}{1.2600E-02} &  \cr
    (n = 1000) & (6.63E+01) & (6.04E+01) & \textcolor{red}{(1.41E+02)}
    & \textcolor{red}{(4.09E+00)} & $\backslash$  \cr
    & & & \textcolor{red}{(failed)} & \textcolor{red}{(failed)} & \cr
\hline
  \end{tabular}}
\end{table}

\vskip 2mm

\begin{table}[htbp]
  \renewcommand{\arraystretch}{1.5}
  \centering
  \caption{Numerical results of CUTEst problems \cite{GOT2015} (no.101-116) computed
  by CNMGE, MCS, GLODS, and VRBBO.}
  \label{TABCOMCUTE8}
  \resizebox{\textwidth}{!}{
    \begin{tabular}{|c|c|c|c|c|c|c|c|c|}
    \hline
    &
    \multicolumn{1}{c|}{CNMGE} & \multicolumn{1}{c|}{MCS} & \multicolumn{1}{c|}{GLODS}
    & \multicolumn{1}{c|}{VRBBO} &  \cr \cline{2-5}
    Problem & $\min f(x)$  & $\min f(x)$  & $\min f(x)$   &  $\min f(x)$ & Global minimum \cr
    & (CPU time (s)) & (CPU time (s)) &  (CPU time (s)) & (CPU time (s)) &  \cr \hline
    101. chainwoo
    & 1 & \textcolor{red}{1.8866E+03} & \textcolor{red}{2.0928E+04}
    & \textcolor{red}{1.1987E+03} &  \cr
    (n = 1000) & (9.19E+01) & \textcolor{red}{(1.05E+02)} & \textcolor{red}{(1.43E+02)}
    & \textcolor{red}{(5.12E+00)} & $\backslash$  \cr
    &  & \textcolor{red}{(failed)} & \textcolor{red}{(failed)} & \textcolor{red}{(failed)}
    & \cr \hline
    102. chnrosnb
    & 0 & 5.4636E-06 & \textcolor{red}{4.0289E+01} & 1.1332E-04 &  \cr
    (n = 50) & (2.29E+00) & (6.52E+00) & \textcolor{red}{(6.80E+00)}
    & (7.85E-01) & $\backslash$ \cr
\hline
  \end{tabular}}
\end{table}

\vskip 2mm

\begin{table}[htbp]
  \renewcommand{\arraystretch}{1.5}
  \centering
  \caption{Numerical results of CUTEst problems \cite{GOT2015} (no.117-132) computed
  by CNMGE, MCS, GLODS, and VRBBO.}
  \label{TABCOMCUTE9}
  \resizebox{\textwidth}{!}{
    \begin{tabular}{|c|c|c|c|c|c|c|c|c|}
    \hline
    &
    \multicolumn{1}{c|}{CNMGE} & \multicolumn{1}{c|}{MCS} & \multicolumn{1}{c|}{GLODS}
    & \multicolumn{1}{c|}{VRBBO} &  \cr \cline{2-5}
    Problem & $\min f(x)$  & $\min f(x)$  & $\min f(x)$   &  $\min f(x)$ & Global minimum \cr
    & (CPU time (s)) & (CPU time (s)) &  (CPU time (s)) & (CPU time (s)) &  \cr \hline
    117. hilbertb
    & 0 & 0 & 0 & \textcolor{red}{1.64E-01} &  \cr
    (n = 50) & (4.14E-01) & (4.44E+00) & (3.93EE+00) & \textcolor{red}{(1.54E+001)}
    & $\backslash$  \cr
    &  &  &  & \textcolor{red}{(failed)}  & \cr \hline
    118. himmelbh
    & -1.0000E+10 & \textcolor{red}{-1.0000E+00} & -1.0000E+96
    & \textcolor{red}{-1.0000E+00} &  \cr
    (n = 2) & (1.75E-01) & \textcolor{red}{(1.60E-01)}
    & (6.16E+01) & \textcolor{red}{(3.19E-02)} & $\backslash$ \cr
\hline
  \end{tabular}}
\end{table}

\vskip 2mm

\begin{table}[htbp]
  \renewcommand{\arraystretch}{1.5}
  \centering
  \caption{Numerical results of CUTEst problems \cite{GOT2015} (no.133-148) computed
  by CNMGE, MCS, GLODS, and VRBBO.}
  \label{TABCOMCUTE10}
  \resizebox{\textwidth}{!}{
    \begin{tabular}{|c|c|c|c|c|c|c|c|c|}
    \hline
    &
    \multicolumn{1}{c|}{CNMGE} & \multicolumn{1}{c|}{MCS} & \multicolumn{1}{c|}{GLODS}
    & \multicolumn{1}{c|}{VRBBO} &  \cr \cline{2-5}
    Problem & $\min f(x)$  & $\min f(x)$  & $\min f(x)$   &  $\min f(x)$ & Global minimum \cr
    & (CPU time (s)) & (CPU time (s)) &  (CPU time (s)) & (CPU time (s)) &  \cr \hline
    133. vardim
    & 0 & \textcolor{red}{7.8388E+04} & \textcolor{red}{2.0284E+07} &
    \textcolor{red}{2.7162E+05} &  \cr
    (n = 100) & (1.46E-01) & \textcolor{red}{(8.78E+01)} &
    \textcolor{red}{(9.38E+00)} & \textcolor{red}{(3.75E-01)} & $\backslash$  \cr
    & & \textcolor{red}{(failed)} & \textcolor{red}{(failed)} &
    \textcolor{red}{(failed)} & \cr \hline
    134. watson
    & 1.8000E-07 & 5.6003E-07 & 5.6725E-04 & 5.0000E-03 &  \cr
    (n = 31) & (7.18E+00) & (1.20E+01) & (8.04E+00) & (1.80E+00) & $\backslash$
    \cr
\hline
  \end{tabular}}
\end{table}

\end{appendix}

\end{document}